\documentclass[conference]{IEEEtran}
\PassOptionsToPackage{numbers,compress}{natbib}
\usepackage[numbers]{natbib}
\usepackage[title]{appendix}
\usepackage{multicol}
\usepackage{bm}
\usepackage{times}

\usepackage{graphicx}
\usepackage{amsmath}
\usepackage{amssymb}
\usepackage{booktabs}
\usepackage{cuted}
\usepackage{duckuments}
\usepackage{subcaption}

\usepackage{appendix}
\usepackage{titletoc}

\usepackage[table]{xcolor}
\definecolor{mycitecolor}{RGB}{0, 128, 255} 
\usepackage[pagebackref,breaklinks,colorlinks, citecolor=mycitecolor]{hyperref}

\usepackage[capitalize]{cleveref}
\Crefname{section}{Sec.}{Secs.}
\crefname{section}{Sec.}{Secs.}
\Crefname{table}{Tab.}{Tabs.}
\crefname{table}{Tab.}{Tabs.}
\Crefname{figure}{Fig.}{Figs.}
\crefname{figure}{Fig.}{Figs.}

\usepackage[utf8]{inputenc} 
\usepackage[T1]{fontenc}    
\usepackage{url}            
\usepackage{amsfonts}       
\usepackage{nicefrac}       
\usepackage[nopatch=eqnum]{microtype}      

\usepackage{epsfig}
\usepackage{tabularx}
\usepackage{bbm}
\usepackage{color}
\usepackage{wrapfig}
\usepackage{subcaption}
\usepackage{float}
\usepackage{makecell}
\usepackage[percent]{overpic}
\usepackage{adjustbox}

\usepackage{multirow}
\usepackage{inconsolata}
\usepackage{comment}
\usepackage{xspace}
\usepackage{utfsym}

\usepackage[font=small]{caption}

\newcommand{\GG}{\mathcal{G}}
\newcommand{\VV}{\mathcal{V}}
\newcommand{\EE}{\mathcal{E}}

\newcolumntype{Y}{>{\centering\arraybackslash}X}

\makeatletter
\def\blfootnote{\gdef\@thefnmark{}\@footnotetext}
\makeatother

\usepackage{soul}
\usepackage{siunitx}
\usepackage{dsfont}
\usepackage{hyperref}
\newcounter{mysection}

\begin{document}

\title{Particle-Grid Neural Dynamics for Learning Deformable Object Models from RGB-D Videos
}

\author{
\authorblockN{Kaifeng Zhang\textsuperscript{1}\quad 
Baoyu Li\textsuperscript{2}\quad  
Kris Hauser\textsuperscript{2}\quad
Yunzhu Li\textsuperscript{1}}
\authorblockN{
\textsuperscript{1}Columbia University\quad
\textsuperscript{2}University of Illinois Urbana-Champaign}
}

\maketitle
\IEEEpeerreviewmaketitle

\setcounter{figure}{0}
\setcounter{table}{0}
\setcounter{section}{0}
\renewcommand{\thesection}{\Roman{section}}
\renewcommand*{\theHsection}{\thesection} 

\begin{strip}
    \centering
    \vspace{-30pt}
    \includegraphics[width=\linewidth]{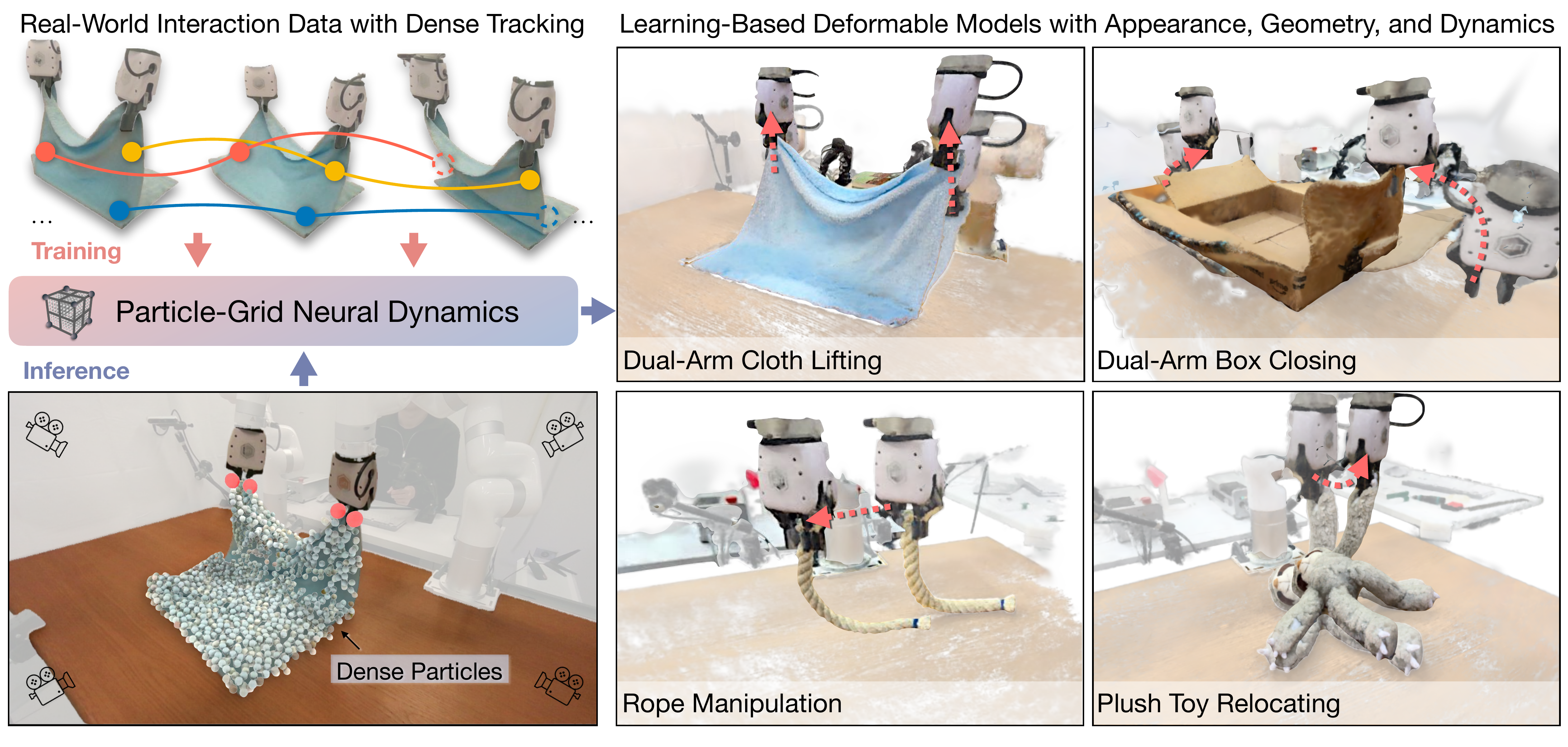}
    \vspace{-20pt}
    \captionof{figure}{\small
    Modeling deformable objects from RGB-D videos presents a significant challenge due to occlusions and complex physical interactions.
    Our \textbf{Particle-Grid Neural Dynamics} framework learns the behavior of deformable objects directly from real-world observations. To train the model, we introduce a novel dense 3D tracking method that leverages foundational vision models for video tracking. The trained model predicts the motion of dense particles under robot-object interactions. We demonstrate the ability of Particle-Grid Neural Dynamics to model complex interactions across a diverse set of objects, including ropes, cloth, plush toy, box, and bread.
    }
    \label{fig:teaser}
    \vspace{-7pt}
\end{strip}
\begin{abstract}
Modeling the dynamics of deformable objects is challenging due to their diverse physical properties and the difficulty of estimating states from limited visual information. We address these challenges with a neural dynamics framework that combines object particles and spatial grids in a hybrid representation. Our particle-grid model captures global shape and motion information while predicting dense particle movements, enabling the modeling of objects with varied shapes and materials. Particles represent object shapes, while the spatial grid discretizes the 3D space to ensure spatial continuity and enhance learning efficiency. Coupled with Gaussian Splattings for visual rendering, our framework achieves a fully learning-based digital twin of deformable objects and generates 3D action-conditioned videos. Through experiments, we demonstrate that our model learns the dynamics of diverse objects--such as ropes, cloths, stuffed animals, and paper bags--from sparse-view RGB-D recordings of robot-object interactions, while also generalizing at the category level to unseen instances. Our approach outperforms state-of-the-art learning-based and physics-based simulators, particularly in scenarios with limited camera views. Furthermore, we showcase the utility of our learned models in model-based planning, enabling goal-conditioned object manipulation across a range of tasks. The project page is available at \texttt{\url{https://kywind.github.io/pgnd}}.
\end{abstract}

\section{Introduction}

Learning predictive models is crucial for a wide range of robotic tasks. In deformable object manipulation, an accurate predictive object dynamics model enables model-based planning, policy evaluation, and real-to-sim asset generation. However, developing dynamics models for deformable objects that are both accurate and generalizable remains a significant challenge. For example, physics-based simulators~\cite{hu2018moving, pbd2014} often struggle to generalize to the real world due to the inherent sim-to-real gap and the difficulties of system identification and state estimation. Meanwhile, video-based predictive models~\cite{finn2017deep, yang2023learning} are computationally expensive, lack 3D spatial understanding, and are highly sensitive to viewpoint and appearance changes.

Recent advancements in deformable object modeling have focused on learning dynamics models for particles that encode 3D geometric information directly obtained from RGB-D cameras. The current state-of-the-art methods in this area utilize Graph Neural Networks (GNNs), where particle sets are modeled as spatial adjacency graphs, and message passing is performed to aggregate information and generate motion predictions~\cite{whitney2023learning3dparticlebasedsimulators, zhang2024dynamic}. However, these methods face significant challenges: the effectiveness of message passing is highly sensitive to the spatial distribution and connectivity of the graph nodes, making them vulnerable to partial observations. Additionally, insufficient message-passing steps may fail to capture global information, while excessive steps can lead to overly smoothed predictions. As a result, these approaches are often limited to relatively simple simulated environments or real-world objects with easily perceivable geometries, where graph construction is straightforward with nearest neighbors.

To address these limitations, we introduce a novel class of dynamic models called particle-grid neural dynamics. This model leverages a hybrid representation that combines object particles with fixed spatial grids. It takes the kinematic states of the particles as input and predicts a spatial velocity field at fixed grid points. A global point cloud encoder is employed to capture comprehensive information from the particle set, effectively overcoming the connectivity challenges commonly faced by GNNs and enhancing the model’s robustness to incomplete observations. Additionally, this encoder allows for the processing of denser particle sets compared to sparse graph nodes, thereby enriching the model with finer geometric details. The grid representation further acts as a regularizer, ensuring spatial continuity in velocity predictions while also reducing computational costs when handling large numbers of particles. By combining object particles with spatial grids, our framework parameterizes dynamics in both Lagrangian and Eulerian coordinates, drawing an analogy to physics-based deformable object simulation methods~\cite{sulsky1995application, hu2018moving}. By utilizing neural networks as message integrators to bridge particle and grid representations, our model handles a diverse range of materials without requiring material-specific physical laws, and can operate effectively with only partial observations as input.

Notably, our model is trained purely on RGB-D videos of robots interacting with objects through diverse and non-task-specific behaviors. To extract object particles from raw recordings, we have developed a 3D fusion and tracking framework. This framework utilizes foundational vision models~\cite{kirillov2023segment, ren2024grounded, karaev24cotracker3} to estimate segmentation masks and pixel tracks, which are then fused in 3D to generate persistent, dense 3D tracks that serve as training data for the neural dynamics model. 

In our experiments, we demonstrate that the proposed particle-grid neural dynamics model can effectively simulate a diverse range of challenging deformable objects, including ropes, cloth, plush toys, boxes, paper bags, and bread. We compare our model with existing state-of-the-art approaches and show that it consistently outperforms the baselines in prediction accuracy. Additionally, we highlight that our dynamics model seamlessly integrates with 3D appearance reconstruction methods such as 3D Gaussian Splatting (3DGS)~\cite{kerbl3Dgaussians}, enabling the fusion of both techniques to generate highly realistic renderings of predictions. This improved accuracy in dynamics prediction directly translates to higher-fidelity renderings compared to existing methods. To further validate its robustness, we conduct experiments that demonstrate the model's effectiveness under sparse-view conditions. Finally, we showcase the model's applicability to deformable object manipulation by incorporating it into a Model Predictive Control (MPC) framework, underscoring its potential for real-world robotic applications.

\label{sec:intro}

\begin{figure*}[t]
    \centering
    \includegraphics[width=\linewidth]{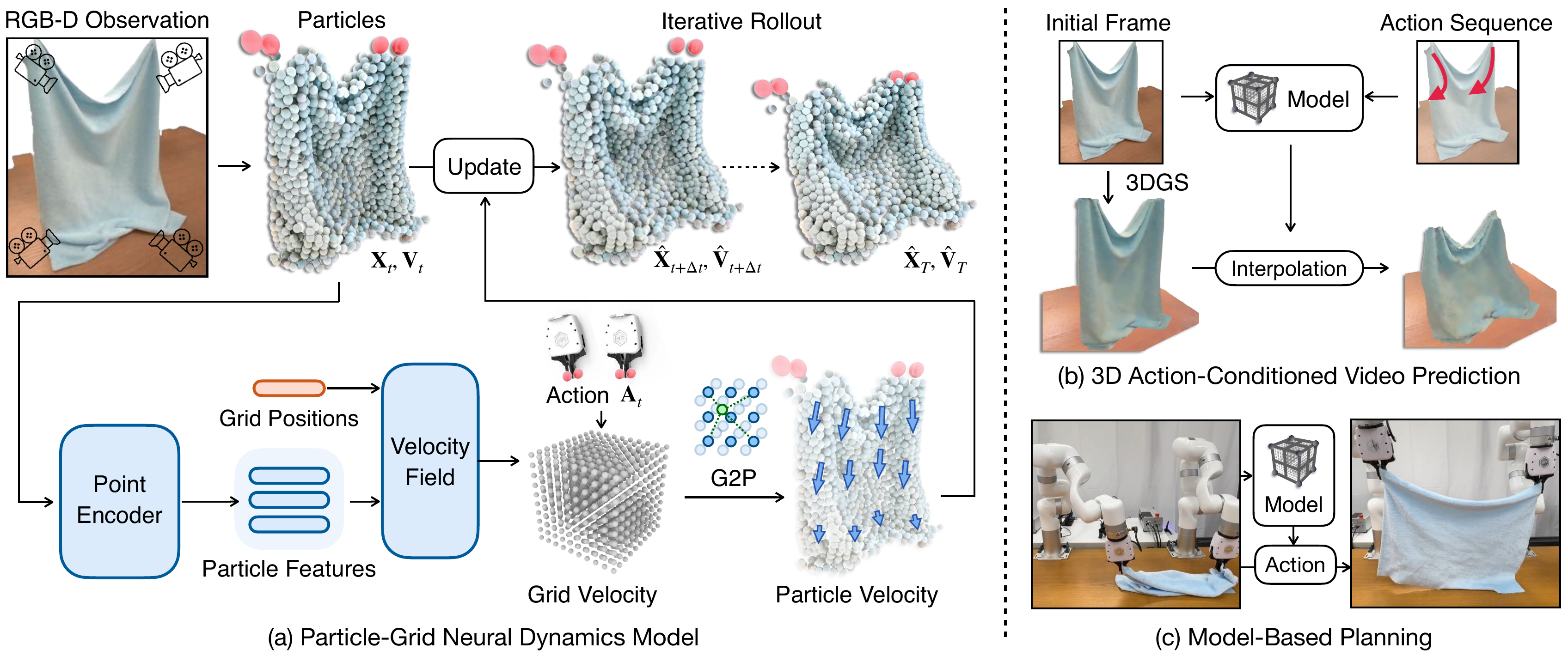}
    \caption{\small
    \textbf{Overview of proposed framework: Particle-Grid Neural Dynamics.} \textbf{(a)}~A diagram of our dynamics model. Given particle positions $\mathbf{X}_t$ and velocities $\mathbf{V}_t$ fused from multi-view depth images as input, our model predicts dense per-particle motion by first using a point encoder to extract particle features and predict the velocity field, which is then transformed into a grid representation to estimate the velocity distribution in 3D space. The model updates particle positions $\hat{\mathbf{X}}_{t+\Delta t}$ with the predicted velocities $\hat{\mathbf{V}}_{t+\Delta t}$ to perform iterative rollouts. \textbf{(b)}~Our framework enables 3D action-conditioned video prediction by reconstructing objects with 3D Gaussian Splatting and interpolating the 6DoF transformation of Gaussian kernels using the predicted particle motions. \textbf{(c)}~The model can be integrated into model-based planning frameworks to generate plausible motions for manipulating deformable objects.
    }
    \label{fig:overview}
    \vspace{-15pt}
\end{figure*}
\section{Related Work}
\subsection{Physics-Based Deformable Modeling.}
The simulation of deformable objects is essential for advancing both the modeling and robotic manipulation of soft, flexible materials. Researchers have introduced various analytical physics-based approaches, including, but not limited to, the mass-spring system~\cite{fastsimmassspring2013, diffclothsim2019}, the Finite Element Method (FEM)~\cite{FEMdeform2006, peng20233dforcecontactestimation, du2021_diffpd}, the Discrete Elastic Rods (DER)~\cite{DER2008, chen2024differentiableDERdlo}, the Position-Based Dynamics (PBD)~\cite{pbd2014, corl2020softgym, pdbrope2023liu}, and the Material Point Method (MPM)~\cite{sulsky1995application, hu2018moving, difftaichi2020, ma2023learning}.
However, real-world analytical deformable modeling remains challenging due to the difficulty in property identification and state estimation.
While our method uses a hybrid particle-grid representation similar to MPM, we leverage neural networks as message integrators and reduce dependence on full-state information as input. This enhances robustness to partial observations and enables the handling of a wide variety of deformable objects without requiring predefined material-specific constitutive laws.
Recently, the emergence of 3D Gaussian Splatting (3DGS) ~\cite{kerbl3Dgaussians, luiten2023dynamic, duisterhof2023md} has enabled the fusion of Gaussian Splatting reconstructions with physics-based models to simulate the dynamics of deformable objects~\cite{xie2023physgaussian, qiu-2024-featuresplatting, springmassgs2024zhong, zhang2024physdreamer, abou-chakra2024physicallyembodiedgs}.
In our particle-grid neural dynamics model, the particle motion predictions can also be integrated with Gaussian Splatting reconstructions, facilitating the simultaneous modeling and rendering of deformable objects, purely learned from real data.

\subsection{Learning-Based Deformable Modeling.}
Learning-based dynamics models, which use deep neural networks to model the future evolution of dynamical systems, have demonstrated effectiveness across various robotic tasks~\cite{hoque2020visuospatial, learningdlo2020, kipf2019contrastive, zhou2024dinowmworldmodelspretrained, xue2023neural, bauer2024doughnet}. 
Among these learning-based methods, Graph-Based Neural Dynamics (GBND) have shown great promise, as they explicitly model spatial relational biases within complex physical systems~\cite{li2018learning, pfaff2020learning, sanchez2020learning, gnnrigidcontactdynamics2023allen, whitney2023learning3dparticlebasedsimulators}. 
Previous research has investigated the use of GBND across a range of material types, such as rigid bodies~\cite{huang2023defgraspnets, liu2023modelbased, ai2024robopack}, plasticine~\cite{shi2022robocraft, shi2023robocook}, fabrics~\cite{lin2022learning, longhini2023edo, puthuveetil2023robust, longhini2024clothsplatting},
ropes~\cite{gnncable2022}, and granular piles~\cite{Wang-RSS-23, shen2024babndlonghorizonmotionplanning}. 
Beyond simulation and single-material scenarios, GBND has also demonstrated flexibility and generalization in modeling diverse materials using a unified framework~\cite{zhang2024adaptigraph, zhang2024dynamic}. 
However, these approaches often operate on spatially sparse graph vertices, rely on expert knowledge to determine graph connectivity, and do not consider partial observations. For learning dense particle dynamics, \citet{whitney2024modelingrealworldhighdensity} propose transformer-based backbones for higher computational efficiency. However, their work mainly focuses on modeling rigid objects for grasping and pushing tasks. In contrast, our work emphasizes deformable object modeling using a hybrid particle-grid neural dynamics framework, achieving dense particle prediction while remaining robust to incomplete observations and flexible for diverse types of deformable objects with distinct physical properties.

\section{Methods}

Our Particle-Grid Neural Dynamics framework models the dynamics of objects represented by a set of particles. The core of this framework is a dynamics function that predicts the future motion of each particle based on its current and historical states, as well as the current action of the robot’s end effector. A detailed description of the model is provided in Section \ref{sec:3.1} and \ref{sec:3.2}. We introduce the data collection pipeline and the model's training method in Section \ref{sec:3.3}. Additionally, we explore the integration of Particle-Grid Neural Dynamics with 3D Gaussian Splatting for 3D video rendering, as discussed in Section \ref{sec:3.4}. The application of this model within a Model Predictive Control (MPC) framework is covered in Section \ref{sec:3.5}. An overview of our method is also provided in Fig.~\ref{fig:overview}.

\subsection{Particle-Grid Neural Dynamics}
\label{sec:3.1}
\subsubsection{State Representation}
We intend to learn a particle-based dynamics model, which represents the target object as a collection of particles $\mathbf{X}_t \in \mathbb{R}^{3\times n}$, where $n$ is the number of particles, and $t$ is the time. The velocity of the particles, $\mathbf{V}_t \in \mathbb{R}^{3\times n}$ is defined as the time derivative of $\mathbf{X}$ at time $t$.

\subsubsection{Action Representation}
The action $\mathbf{A}_t$ represents the external effects caused to the object by the robot at time $t$. We define $\mathbf{A}_t$ as
\begin{equation}
    \mathbf{A}_t = (\mathbf{y}, \mathbf{T}_t, \mathbf{\dot{T}}_t, \mathbf{o}_t),
\end{equation}
where $\mathbf{y}$ is the action type label, $\mathbf{T}_t$ is the end-effector pose, $\mathbf{\dot{T}}_t$ its time derivative, and $\mathbf{o}_t$ the gripper open distance. 
In our experiments $\mathbf{y}$ is a binary label indicating whether the action is a grasped or nonprehensile interaction; the implementation details of these two types are given in Sec.~\ref{sec:GridVelocityEditing}.

\subsubsection{Dynamics Function}
We consider the change of state caused by the state itself (e.g., objects falling due to gravity) and the robot's actions (e.g., robots grasping an object thus making it move) in the object's dynamic functions:
\begin{equation}
    \hat{\mathbf{V}}_{t+\Delta t} = \mathbf{f}(\mathbf{X}_{t}, \mathbf{V}_{t}, \mathbf{A}_t),
\end{equation}
where $\mathbf{f}$ is the dynamics function predicting the state evolution.
Since the particle representation typically cannot capture the full state information (e.g., internal stress or contact mode with other objects), a common practice is to incorporate historical states as additional inputs for making predictions:
\begin{equation}
    \hat{\mathbf{V}}_{t+\Delta t} = \mathbf{f}(\mathbf{X}_{t-h \Delta t:t}, \mathbf{V}_{t-h \Delta t:t}, \mathbf{A}_t),
\end{equation}
where $h$ is the history window size. In our neural dynamics model, we approximate the function $\mathbf{f}$ using a neural network parameterized by $\theta$ and derive the next particle positions by applying forward Euler time integration:
\begin{equation}
    \hat{\mathbf{X}}_{t+\Delta t} = \mathbf{X}_{t} + \Delta t \cdot \mathbf{f}_\theta(\mathbf{X}_{t-h\Delta t:t}, \mathbf{V}_{t-h\Delta t:t}, \mathbf{A}_t).
    \label{eq:time_integration}
\end{equation}

\subsubsection{The Particle-Grid Model}
Learning the neural dynamics parameters $\theta$ with an end-to-end neural network usually leads to unstable predictions due to accumulative error. Motivated by the use of hybrid Lagrangian-Eulerian representations in MPM~\cite{sulsky1995application}, our model also utilizes a hybrid particle-grid representation to inject inductive bias related to spatial continuity and local information integration. 
Specifically, we define a uniformly distributed grid in the space as 
\begin{equation}
    \mathbf{G}_{l_x, l_y, l_z, \delta} = \{(k_x \delta, k_y \delta, k_z \delta) | k_i \in [l_i], \forall i \in \{x, y, z\}\}.
\end{equation}
The parameters $l_x, l_y, l_z, \delta$ control the spatial limits and resolution of the grid. Empirically, we set $l_x, l_y, l_z$ to 100 or 50 and $\delta$ to $\SI{1}{\centi\meter}$ or $\SI{2}{\centi\meter}$ to balance computational cost and resolution. To enforce translational invariance during prediction, we always translate the positions of the particles and the robot end effector to the volume defined by $\mathbf{G}_{l_x, l_y, l_z, \delta}$. 

The particle-grid dynamics function is defined by the following components:
\begin{equation}
    \mathbf{f}_{\theta} = \mathbf{h}^{\text{G2P}} \cdot \mathbf{g}^{\text{grid}} \cdot \mathbf{f}^{\text{field}}_{\psi} \cdot \mathbf{f}^{\text{feature}}_{\phi},
\end{equation}
where $\mathbf{f}^{\text{feature}}_{\phi}$ is the neural network-based point encoder for extracting feature from the input particles (Sec.~\ref{sec:PointEncoder}), $\mathbf{f}^{\text{field}}_{\psi}$ is the neural network-based function for parameterizing a neural velocity field based on the extracted features (Sec.~\ref{sec:NeuralVelocityField}), and $\mathbf{g}^{\text{grid}}$ is a grid velocity editing (GVE) control method to encode collision surfaces and robot gripper movements (Sec.~\ref{sec:GridVelocityEditing}), and $\mathbf{h}^{\text{G2P}}$ is the grid-to-particle integration function for calculating particle velocities from grid-based velocity field (Sec.~\ref{sec:G2P}).

\subsection{Model Components}
\label{sec:3.2}
\subsubsection{Point Encoder}
\label{sec:PointEncoder}
The point encoder encodes particle positions and velocities to per-particle latent features $\mathbf{Z}_t \in \mathbb{R}^{d\times n}$, where $d$ is the feature dimension: 
\begin{equation}
    \mathbf{Z}_t = \mathbf{f}^{\text{feature}}_\phi (\mathbf{X}_{t-h \Delta t:t}, \mathbf{V}_{t-h \Delta t:t}).
\end{equation}
We use PointNet~\cite{qi2017pointnet} as the encoder for its efficiency and strong performance in extracting 3D point features. The encoder captures global information from the set of all particles, including the object's shape and the historical motion of the particles, which are used to implicitly infer the object's physical properties and dynamic state. This is essential for handling incomplete observations, where the feature encoder must extract occlusion-robust features for subsequent velocity decoding.

\subsubsection{Neural Velocity Field}
\label{sec:NeuralVelocityField}
In this step, we use a neural implicit function $\mathbf{f}^{\text{field}}_\psi$ to predict a spatial velocity grid, at time $t$, from the extracted point features. For $g \in \mathbf{G}_{l_x, l_y, l_z, \delta}$, the function $\mathbf{f}^{\text{field}}_\psi$ is instantiated as an MLP that takes the grid locations $\mathbf{x}_{g}$ and the corresponding locality-aware feature $\mathbf{z}_{g, t}$ as inputs, then predict per-grid velocity vector $\mathbf{v}_{g, t}$ by
\begin{equation}
    \hat{\mathbf{v}}_{g, t} = \mathbf{f}^{\text{field}}_\psi (\gamma(\mathbf{x}_{g}), \mathbf{z}_{g, t}), \label{eq:query}
\end{equation}
where $\gamma$ is the sinusoidal positional encoding, and the locality-aware feature $\mathbf{z}_{g, t}$ is defined as the average pooling of particle features within the neighborhood of grid location:
\begin{equation}
    \mathbf{z}_{g, t} = \frac{\sum_{p \in \mathcal{N}_r(\mathbf{X}_t, \mathbf{x}_g)} \mathbf{z}_{p,t}}{|\mathcal{N}_r(\mathbf{X}_t, \mathbf{x}_g)|}, \label{eq:local_feature}
\end{equation}
where $\mathcal{N}_r(\mathbf{X}_t, \mathbf{x}_g)$ is the set of indices of particles within $\mathbf{X}_t$ whose positions are within radius $r$ of the grid location $\mathbf{x}_g$. By incorporating the radius hyperparameter $r$, we can control the number of particle features a grid point attends to, thus encouraging the network to predict velocities that are dependent on local geometry. Empirically, we set $r=\SI{0.2}{\meter}$.

\subsubsection{G2P}
\label{sec:G2P}
After calculating the grid's velocities, we transfer from the Eulerian grid to Lagrangian particles via spline interpolation. Following MPM, we utilize a continuous B-spline kernel to transfer grid velocities $\hat{\mathbf{v}}_{g, t}$ to particle velocities $\hat{\mathbf{v}}_{p, t}$:
\begin{equation}
    \hat{\mathbf{v}}_{p, t} = \sum_{g\in \mathbf{G}} 
    \hat{\mathbf{v}}_{g, t} w_{pg, t}, \label{eq:g2p}
\end{equation}
where the $w_{pg, t}$ is the value of the B-spline kernel defined on the grid position $\mathbf{x}_g$ and evaluated at the particle location $\mathbf{x}_{p, t}$. It assigns larger weights to closer grid-particle pairs, achieving smooth spatial interpolation. The predictions $\hat{\mathbf{V}} \in \mathbb{R}^{3\times n}$ serves as the final output of the dynamics function $\mathbf{f}_\theta$ and is used to perform time integration in Eq.~\ref{eq:time_integration}.

\definecolor{highlightblue}{RGB}{214, 235, 243}  

\begin{table*}[ht]
    \centering
    \footnotesize
    \resizebox{0.97\textwidth}{!}
    {
        \begin{tabular}{@{}cccccccc@{}}
            \toprule
            Method      & Metric   & Cloth   & Rope  & Plush   & Box   & Bag   & Bread   \\
            \midrule
            
            \multicolumn{1}{l|}{MPM~\cite{hu2018moving}}      & \multicolumn{1}{c|}{\multirow{4}{*}{MDE $\downarrow$}} & 
            $0.176_{\pm 0.107}$                                                           & 
            $0.138_{\pm 0.072}$                                                           & 
            $0.163_{\pm 0.148}$                                                           & 
            $-$                                                                           & 
            $0.226_{\pm 0.026}$                                                           & 
            $0.034_{\pm 0.014}$                                                           \\
            
            \multicolumn{1}{l|}{GBND~\cite{zhang2024dynamic}}     & \multicolumn{1}{c|}{}                       & 
            $0.077_{\pm 0.033}$                                                           & 
            $0.062_{\pm 0.025}$                                                           & 
            $0.078_{\pm 0.028}$                                                           & 
            $0.045_{\pm 0.008}$                                                           & 
            $0.030_{\pm 0.011}$                                                           & 
            $0.031_{\pm 0.014}$                                                           \\
            
            \multicolumn{1}{l|}{Particle} & \multicolumn{1}{c|}{}                       & 
            $0.059_{\pm 0.039}$                                                           & 
            $0.061_{\pm 0.051}$                                                           & 
            $0.060_{\pm 0.027}$                                                           & 
            $0.025_{\pm 0.009}$                                                           & 
            $0.021_{\pm 0.016}$                                                           & 
            $0.038_{\pm 0.018}$                                                           \\
            
            \multicolumn{1}{l|}{Ours}     & \multicolumn{1}{c|}{}                       & 
            \colorbox{highlightblue}{$\mathbf{0.045}_{\mathbf{\pm 0.023}}$} & 
            \colorbox{highlightblue}{$\mathbf{0.039}_{\mathbf{\pm 0.032}}$} & 
            \colorbox{highlightblue}{$\mathbf{0.043}_{\mathbf{\pm 0.018}}$} & 
            \colorbox{highlightblue}{$\mathbf{0.022}_{\mathbf{\pm 0.007}}$} & 
            \colorbox{highlightblue}{$\mathbf{0.016}_{\mathbf{\pm 0.005}}$} & 
            \colorbox{highlightblue}{$\mathbf{0.020}_{\mathbf{\pm 0.011}}$} \\ 
            
            \midrule
            
            \multicolumn{1}{l|}{MPM~\cite{hu2018moving}}      & \multicolumn{1}{c|}{\multirow{4}{*}{CD $\downarrow$}}  & 
            $0.156_{\pm 0.091}$                                                           & 
            $0.115_{\pm 0.054}$                                                           & 
            $0.153_{\pm 0.207}$                                                           & 
            $-$                                                                           & 
            $0.183_{\pm 0.032}$                                                           & 
            $0.031_{\pm 0.010}$                                                           \\
            
            \multicolumn{1}{l|}{GBND~\cite{zhang2024dynamic}}     & \multicolumn{1}{c|}{}                       & 
            $0.083_{\pm 0.034}$                                                           & 
            $0.073_{\pm 0.027}$                                                           & 
            $0.064_{\pm 0.016}$                                                           & 
            $0.062_{\pm 0.014}$                                                           & 
            $0.042_{\pm 0.007}$                                                           & 
            $0.031_{\pm 0.013}$                                                           \\
            
            \multicolumn{1}{l|}{Particle} & \multicolumn{1}{c|}{}                       & 
            $0.051_{\pm 0.034}$                                                           & 
            $0.059_{\pm 0.058}$                                                           & 
            $0.043_{\pm 0.019}$                                                           & 
            $0.032_{\pm 0.011}$                                                           & 
            $0.025_{\pm 0.016}$                                                           & 
            $0.044_{\pm 0.031}$                                                           \\
            
            \multicolumn{1}{l|}{Ours}     & \multicolumn{1}{c|}{}                       & 
            \colorbox{highlightblue}{$\mathbf{0.043}_{\mathbf{\pm 0.022}}$} & 
            \colorbox{highlightblue}{$\mathbf{0.038}_{\mathbf{\pm 0.036}}$} & 
            \colorbox{highlightblue}{$\mathbf{0.033}_{\mathbf{\pm 0.013}}$} & 
            \colorbox{highlightblue}{$\mathbf{0.015}_{\mathbf{\pm 0.003}}$} & 
            \colorbox{highlightblue}{$\mathbf{0.021}_{\mathbf{\pm 0.005}}$} & 
            \colorbox{highlightblue}{$\mathbf{0.018}_{\mathbf{\pm 0.012}}$} \\ 
            
            \midrule
            
            \multicolumn{1}{l|}{MPM~\cite{hu2018moving}}      & \multicolumn{1}{c|}{\multirow{4}{*}{EMD $\downarrow$}} & 
            $0.093_{\pm 0.078}$                                                           & 
            $0.081_{\pm 0.052}$                                                           & 
            $0.092_{\pm 0.131}$                                                           & 
            $-$                                                                           & 
            $0.110_{\pm 0.027}$                                                           & 
            $0.021_{\pm 0.009}$                                                           \\
            
            \multicolumn{1}{l|}{GBND~\cite{zhang2024dynamic}}     & \multicolumn{1}{c|}{}                       & 
            $0.035_{\pm 0.017}$                                                           & 
            $0.036_{\pm 0.016}$                                                           & 
            $0.032_{\pm 0.010}$                                                           & 
            $0.032_{\pm 0.008}$                                                           & 
            $0.016_{\pm 0.005}$                                                           & 
            $0.016_{\pm 0.008}$                                                           \\
            
            \multicolumn{1}{l|}{Particle} & \multicolumn{1}{c|}{}                       & 
            $0.029_{\pm 0.024}$                                                           & 
            $0.036_{\pm 0.037}$                                                           & 
            $0.025_{\pm 0.013}$                                                           & 
            $0.018_{\pm 0.006}$                                                           & 
            $0.012_{\pm 0.010}$                                                           & 
            $0.023_{\pm 0.016}$                                                           \\
            
            \multicolumn{1}{l|}{Ours}     & \multicolumn{1}{c|}{}                       & 
            \colorbox{highlightblue}{$\mathbf{0.022}_{\mathbf{\pm 0.013}}$} & 
            \colorbox{highlightblue}{$\mathbf{0.021}_{\mathbf{\pm 0.021}}$} & 
            \colorbox{highlightblue}{$\mathbf{0.017}_{\mathbf{\pm 0.007}}$} & 
            \colorbox{highlightblue}{$\mathbf{0.016}_{\mathbf{\pm 0.005}}$} & 
            \colorbox{highlightblue}{$\mathbf{0.009}_{\mathbf{\pm 0.003}}$} & 
            \colorbox{highlightblue}{$\mathbf{0.010}_{\mathbf{\pm 0.008}}$} \\ 
            
            \bottomrule
        \end{tabular}
    }
    \caption{
        \textbf{Quantitative Results on Dynamics Prediction.} We compare our method with the Material Point Method (MPM)~\cite{hu2018moving}, Graph-Based Neural Dynamics (GBND)~\cite{zhang2024dynamic}, and a particle-based dynamics model without the grid representation. We report the mean and standard deviation of the prediction error over a 3-second future horizon. The best results are highlighted in bold and blue.
    }
    \label{tab:dynamics_particle}
    \vspace{-10pt}
\end{table*}

\subsubsection{Controlling Deformation}
\label{sec:GridVelocityEditing}

We present two methods for controlling deformations by interactions with external objects: Grid Velocity Editing (GVE) and Robot Particles (RP).
GVE is inspired from MPM approaches and we use it for grasped interactions and object-ground interaction. Simply put, the operator $\mathbf{g}^{\text{grid}}$ changes the velocities on the grid to match physical constraints. For ground contact, we project velocity back from the contact surface and incorporate friction terms. To define the motion of a rigidly grasped point, we calculate the set of grid points $\mathbf{G}_{\text{grasp}, t}$ within a distance $a$ of the grasp center point $\mathbf{x}_{\text{grasp}, t}$. For each point $g \in \mathbf{G}_{\text{grasp}, t}$, we modify the velocities as follows:
\begin{equation}
    \mathbf{v}_{g, t} = \mathbf{\omega}_t \times (\mathbf{x}_g - \mathbf{x}_{\text{grasp}, t}) + \dot{\mathbf{x}}_{\text{grasp}, t},
\end{equation}
where $\mathbf{\omega}_t$ and $\dot{\mathbf{x}}_{\text{grasp}, t}$ are the angular and linear velocities of the gripper at time $t$, and $\mathbf{x}_{g}$ and $\mathbf{v}_{g, t}$ are the position and velocity of grid point $g$, respectively.

The Robot Particles method allows us to model nonprehensile actions for the Box example in which the object is pushed. Here, we represent the robot gripper with additional particles that carry gripper action information, and fuse this into the object point cloud. Specifically, at each step, we augment the point cloud by
\begin{align}
    \mathbf{X}^{\text{aug}}_{t} &= \mathbf{X}_t \cup \mathbf{X}_{\text{robot}, t}, \\
    \mathbf{V}^{\text{aug}}_{t} &= \mathbf{V}_t \cup \mathbf{V}_{\text{robot}, t},
\end{align}
and model the particle-grid dynamics function on the augmented point cloud. This injects action information into particle features but does not explicitly force particles to move at a prescribed velocity, thus supporting nonprehensile manipulation. Our implementation samples points from the gripper shape and calculates their velocities based on the end-effector transformation from proprioception.

\subsection{Data Collection and Training}
\label{sec:3.3}
We collect training data through teleoperation and automatic annotation using foundation models. Specifically, we record multi-view RGB-D videos of random robot-object interactions. For each camera view, we apply Segment-Anything~\cite{kirillov2023segment, ren2024grounded} to extract persistent object masks across the video. The segmented objects are then cropped and tracked over time using CoTracker~\cite{karaev24cotracker3}, providing 2D trajectories. Using depth information, we perform inverse projection to map these 2D velocities into 3D, resulting in multi-view fused point clouds with persistent particle tracking.

With the collected tracking data, we define particle sets and their trajectories over a look-forward time window as $\mathbf{X}_{t-h\Delta t:t+K\Delta t} \in \mathbb{R}^{3\times n\times T}$, alongside corresponding robot actions $\mathbf{A}_{t-h\Delta t:t+K\Delta t}$, where $K$ is the horizon length hyperparameter. Empirically, we set $h=2$ and $K=5$. Model training begins from a given point cloud at time $t$, followed by iterative dynamics model rollouts for $K$ steps.
Since the dynamics function $\mathbf{f}_\theta$ is fully differentiable, we optimize the network parameters $\phi$ and $\psi$ using gradient descent. The loss function is defined as the mean squared error (MSE) between the predicted and actual particle positions:
\begin{equation}
    \mathcal{L} = \sum_{i=1}^K \|\mathbf{\hat{X}}_{t+i\Delta t} - \mathbf{X}_{t+i\Delta t}\|_2^2, \label{eq:loss}
\end{equation}
where $\mathbf{\hat{X}}_{t+i\Delta t}$ is the predicted particle positions at step $i$.

\subsection{Rendering and Action-Conditioned Video Prediction}
\label{sec:3.4}
Our predictions can be integrated with 3D Gaussian Splatting (3DGS) to achieve a realistic rendering of the results. The 3DGS reconstruction of the object is defined as
\begin{equation}
    \mathcal{G} = \{\mathbf{X}_{\text{GS}}, \mathbf{C}, \mathbf{R}_{\text{GS}}, \mathbf{S}, \mathbf{O}\},
\end{equation}
where $\mathbf{X}_{\text{GS}}$, $\mathbf{C}$, $\mathbf{R}_{\text{GS}}$, $\mathbf{S}$, and $\mathbf{O}$ represent the Gaussian kernels' center location, color, rotation, scale, and opacity, respectively. To transform Gaussians between frames, we first apply the dynamics model to the point cloud set $\mathbf{X}$, yielding the next-frame prediction $\mathbf{\hat{X}}$. The points $\mathbf{\hat{X}}$ can either be sampled from $\mathbf{X}_{\text{GS}}$ or obtained from additional point cloud observations within the same coordinate frame with $\mathbf{X}_{\text{GS}}$. The 6-DoF motions of the Gaussian kernels are interpolated using Linear Blend Skinning (LBS)~\cite{10.1145/1276377.1276478}, which updates $\mathbf{X}_{\text{GS}}$ and $\mathbf{R}_{\text{GS}}$ by treating $\mathbf{X}$ as control points and interpolating their predicted motion to generate new Gaussian centers and rotations. We assume that the color, scale, and opacity of the Gaussian splatting remain constant.

\begin{figure*}[t]
    \centering
    \includegraphics[width=\linewidth]{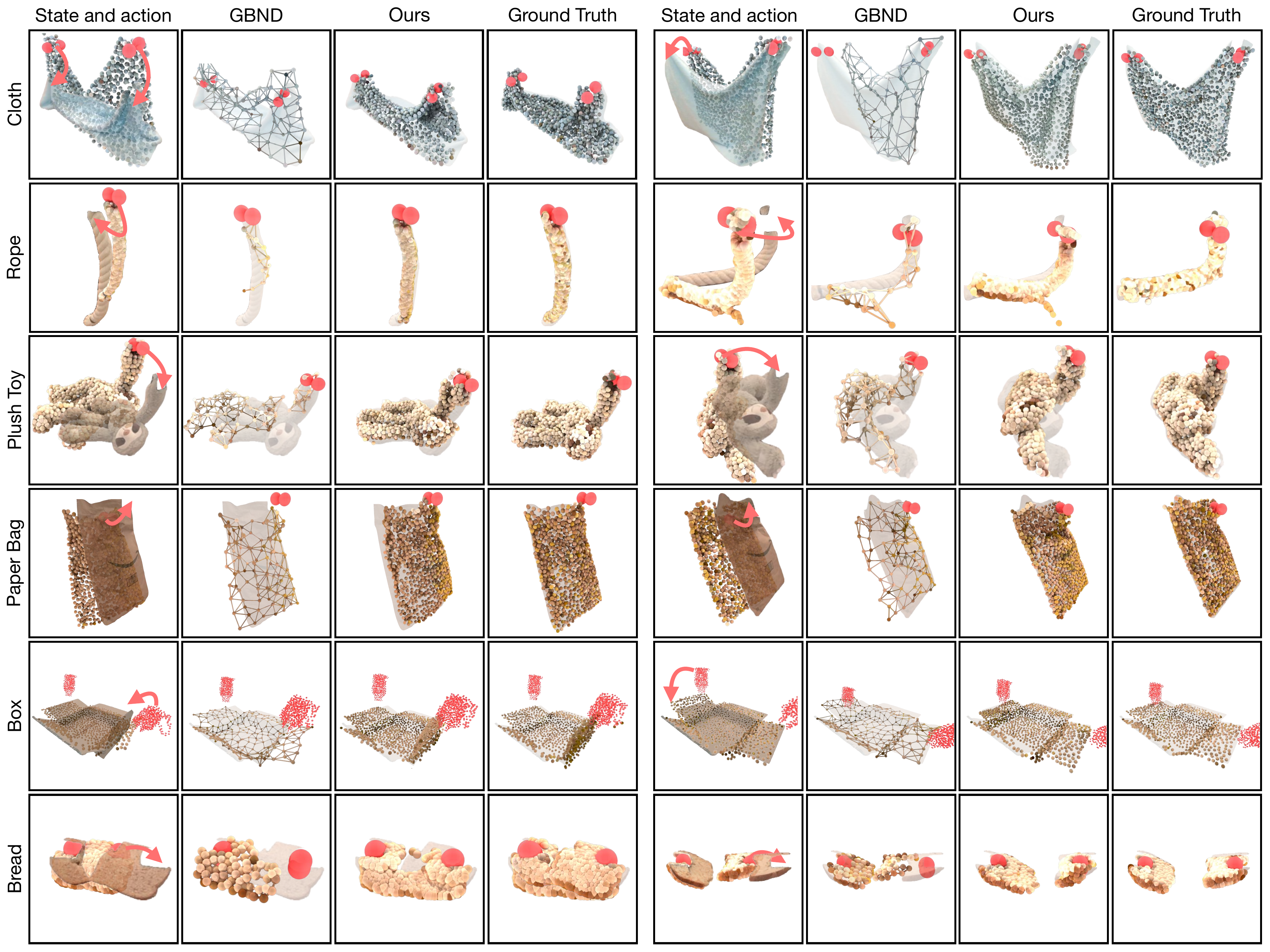}
    \caption{\small
    \textbf{Qualitative Comparisons on Dynamics Prediction.} Given initial states and actions, we show the prediction results of the GBND baseline compared to our particle-grid neural dynamics model. The red spheres indicate the position and orientation of robot grippers. We overlay the predictions with ground truth final state images to highlight the prediction errors. Our model's predictions are more aligned with the ground truth, offering higher-density particle predictions and fewer artifacts compared to the baseline.
    }
    \label{fig:qual_rollout}
    \vspace{-15pt}
\end{figure*}

\subsection{Planning}
\label{sec:3.5}
Our model can be integrated with Model Predictive Control (MPC) for model-based planning. Given multi-view RGB-D captures, we obtain object particles through segmentation, inverse projection into 3D space, and downsampling. The downsampled particles serve as inputs to the dynamics model for future prediction. With a specified cost function, the MPC framework rolls out the dynamics model using sampled actions and optimizes the total cost.
In our experiments, we use the Chamfer Distance between the predicted state $\hat{\mathbf{X}}$ and the target state $\mathbf{X}_{\text{target}}$ as the cost function:
\begin{align}
    \mathbf{J} (\hat{\mathbf{X}}_{1:N}, \mathbf{A}_{1:N}) &= \sum_{t=1}^{N} \text{CD}(\hat{\mathbf{X}}_t, \mathbf{X}_{\text{target}}).  \label{eq:planning} 
\end{align}
We apply the Model-Predictive Path Integral (MPPI)~\cite{williams2017model} trajectory optimization algorithm to minimize the cost and to synthesize the robot's actions. During deployment, we perform online, iterative planning to achieve closed-loop control.

\section{Experiments}
Our experiments are designed to address the following questions:
\begin{itemize}
    \item How well does the particle-grid model learn the dynamics of various types of deformable objects?
    \item Does the model perform effectively under limited visual observation (e.g., sparse views)?
    \item Can we train a unified model for multiple instances within an object category, and how well does it generalize to unseen instances?
    \item Can the model improve the performance of 3D action-conditioned video prediction and model-based planning?
\end{itemize}
We evaluate our method on a diverse set of challenging deformable objects, including cloth, rope, plush toys, bags, boxes, and bread. Our results demonstrate that it outperforms previous state-of-the-art approaches in dynamics prediction accuracy while remaining robust to incomplete camera views. Additionally, we validate the model's capability in category-level training and its effectiveness in downstream applications, such as video prediction and planning.

\begin{figure}[t]
    \centering
    \vspace{-5pt}
    \includegraphics[width=\linewidth]{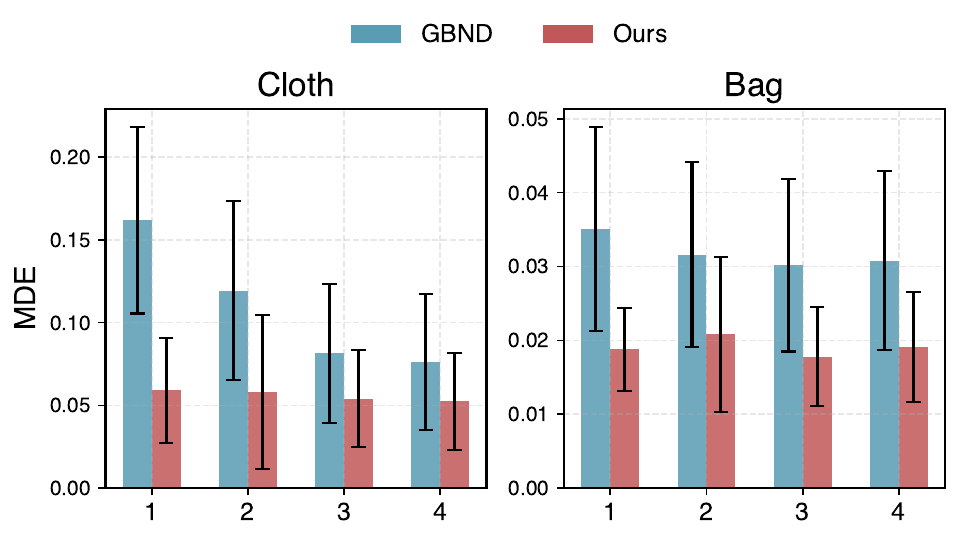}
    \caption{\small
    \textbf{Quantitative Comparisons on Prediction under Partial Views.} We compare our method with the GBND baseline in the cloth and paper bag categories while varying the number of input camera views. We report the mean and standard deviation of the dynamics prediction error. Our method consistently achieves lower error than the baseline, and its error increase rate as the number of camera views decreases is also lower.
    }
    \label{fig:partial}
    \vspace{-15pt}
\end{figure}
\begin{figure}[t]
    \centering
    \includegraphics[width=\linewidth]{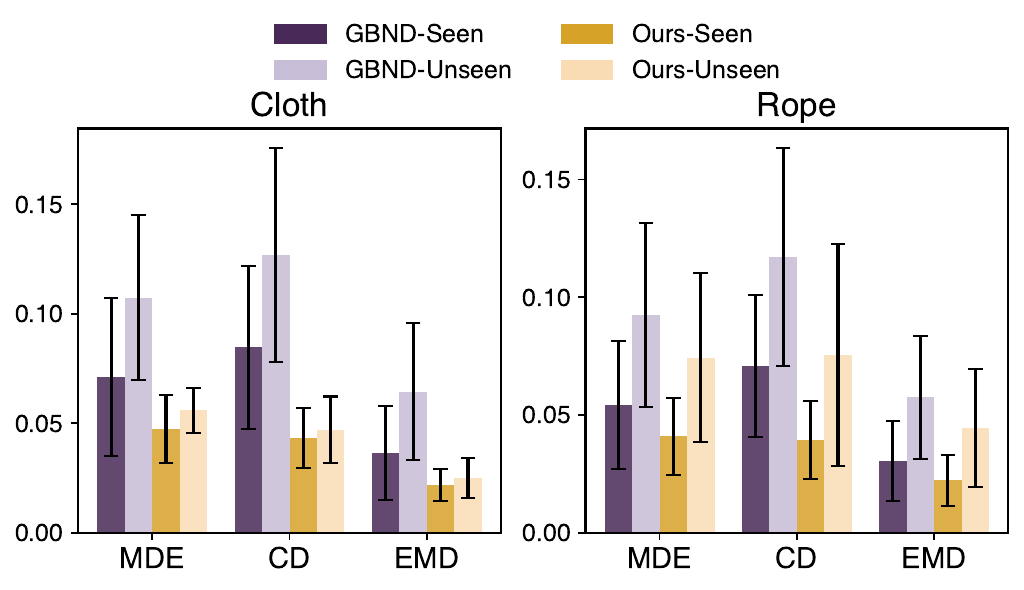}
    \caption{\small
    \textbf{Quantitative Comparisons on Generalization.} Our method is compared with GBND on seen and unseen instances of the rope and cloth categories. We present the mean and standard deviation of dynamics prediction error. Our method's prediction error is lower on both seen and unseen instances compared to the baseline.
    }
    \vspace{-15pt}
    \label{fig:gen}
\end{figure}
\definecolor{highlightblue}{RGB}{214, 235, 243}  

\begin{table*}[ht]
    \centering
    \footnotesize
    \resizebox{0.97\textwidth}{!}{
        \begin{tabular}{@{}cccccccc@{}}
            \toprule
            Method      & Metric   & Cloth   & Rope  & Plush   & Box   & Bag   & Bread   \\ 
            \midrule
            
            \multicolumn{1}{l|}{MPM~\cite{hu2018moving}}      & \multicolumn{1}{c|}{\multirow{4}{*}{$\mathcal{J}$-Score / IoU $\uparrow$}} & 
            $40.5_{\pm 20.3}$                                                           & 
            $12.5_{\pm 9.5}$                                                           & 
            $37.5_{\pm 14.8}$                                                           & 
            $-$                                                                           & 
            $34.3_{\pm 5.5}$                                                           & 
            $42.9_{\pm 10.3}$                                                           \\
            
            \multicolumn{1}{l|}{GBND~\cite{zhang2024dynamic}}     & \multicolumn{1}{c|}{}                       & 
            $56.7_{\pm 13.9}$                                                           & 
            $19.2_{\pm 15.6}$                                                           & 
            $42.7_{\pm 10.5}$                                                           & 
            \colorbox{highlightblue}{$\mathbf{83.0}_{\mathbf{\pm 5.5}}$}                                                           & 
            \colorbox{highlightblue}{$\mathbf{78.0}_{\mathbf{\pm 82.8}}$}                                                           & 
            $49.8_{\pm 13.9}$                                                           \\
            
            \multicolumn{1}{l|}{Particle} & \multicolumn{1}{c|}{}                       & 
            $58.5_{\pm 18.7}$                                                           & 
            $24.4_{\pm 17.7}$                                                           & 
            $49.0_{\pm 12.5}$                                                           & 
            $82.2_{\pm 6.1}$                                                           & 
            $74.5_{\pm 11.7}$                                                           & 
            $39.3_{\pm 15.4}$                                                           \\
            
            \multicolumn{1}{l|}{Ours}     & \multicolumn{1}{c|}{}                       & 
            \colorbox{highlightblue}{$\mathbf{63.2}_{\mathbf{\pm 16.4}}$} & 
            \colorbox{highlightblue}{$\mathbf{29.5}_{\mathbf{\pm 16.5}}$} & 
            \colorbox{highlightblue}{$\mathbf{59.7}_{\mathbf{\pm 11.1}}$} & 
            $82.4_{\pm 5.0}$        & 
            $76.8_{\pm 8.8}$        & 
            \colorbox{highlightblue}{$\mathbf{55.8}_{\mathbf{\pm 10.1}}$} \\ 
            
            \midrule
            
            \multicolumn{1}{l|}{MPM~\cite{hu2018moving}}      & \multicolumn{1}{c|}{\multirow{4}{*}{$\mathcal{F}$-Score $\uparrow$} }  & 
            $28.0_{\pm 11.3}$                                                           & 
            $37.0_{\pm 14.2}$                                                           & 
            $40.4_{\pm 12.2}$                                                           & 
            $-$                                                                           & 
            $13.1_{\pm 5.9}$                                                           & 
            $54.3_{\pm 12.1}$                                                           \\
            
            \multicolumn{1}{l|}{GBND~\cite{zhang2024dynamic}}     & \multicolumn{1}{c|}{}                       & 
            $32.2_{\pm 18.2}$                                                           & 
            $41.9_{\pm 18.9}$                                                           & 
            $35.8_{\pm 11.2}$                                                           & 
            \colorbox{highlightblue}{$\mathbf{74.0}_{\mathbf{\pm 8.8}}$}                                                           & 
            $54.6_{\pm 13.4}$                                                           & 
            $60.1_{\pm 16.1}$                                                           \\
            
            \multicolumn{1}{l|}{Particle} & \multicolumn{1}{c|}{}                       & 
            $41.0_{\pm 19.6}$                                                           & 
            $45.4_{\pm 19.5}$                                                           & 
            $43.0_{\pm 14.0}$                                                           & 
            $73.1_{\pm 8.8}$                                                           & 
            $52.7_{\pm 19.5}$                                                           & 
            $46.8_{\pm 17.7}$                                                           \\
            
            \multicolumn{1}{l|}{Ours}     & \multicolumn{1}{c|}{}                       & 
            \colorbox{highlightblue}{$\mathbf{42.6}_{\mathbf{\pm 20.4}}$} & 
            \colorbox{highlightblue}{$\mathbf{52.6}_{\mathbf{\pm 17.6}}$} & 
            \colorbox{highlightblue}{$\mathbf{53.8}_{\mathbf{\pm 12.9}}$} & 
            $68.8_{\pm 12.9}$       & 
            \colorbox{highlightblue}{$\mathbf{60.3}_{\mathbf{\pm 14.7}}$} & 
            \colorbox{highlightblue}{$\mathbf{64.6}_{\mathbf{\pm 12.7}}$} \\ 
            
            \midrule
            
            \multicolumn{1}{l|}{MPM~\cite{hu2018moving}}      & \multicolumn{1}{c|}{\multirow{4}{*}{LPIPS $\downarrow$}} & 
            $0.141_{\pm 0.060}$                                                           & 
            $0.052_{\pm 0.018}$                                                           & 
            $0.103_{\pm 0.085}$                                                           & 
            $-$                                                                           & 
            $0.145_{\pm 0.020}$                                                           & 
            $0.059_{\pm 0.020}$                                                           \\
            
            \multicolumn{1}{l|}{GBND~\cite{zhang2024dynamic}}     & \multicolumn{1}{c|}{}                       & 
            $0.120_{\pm 0.055}$                                                           & 
            $0.042_{\pm 0.018}$                                                           & 
            $0.081_{\pm 0.023}$                                                           & 
            $0.106_{\pm 0.039}$                                                           & 
            $0.097_{\pm 0.019}$                                                           & 
            $0.052_{\pm 0.015}$                                                           \\
            
            \multicolumn{1}{l|}{Particle} & \multicolumn{1}{c|}{}                       & 
            $0.109_{\pm 0.048}$                                                           & 
            $0.044_{\pm 0.032}$                                                           & 
            $0.072_{\pm 0.024}$                                                           & 
            $0.082_{\pm 0.031}$                                                           & 
            $0.069_{\pm 0.019}$                                                           & 
            $0.054_{\pm 0.020}$                                                           \\
            
            \multicolumn{1}{l|}{Ours}     & \multicolumn{1}{c|}{}                       & 
            \colorbox{highlightblue}{$\mathbf{0.099}_{\mathbf{\pm 0.044}}$} & 
            \colorbox{highlightblue}{$\mathbf{0.041}_{\mathbf{\pm 0.033}}$} & 
            \colorbox{highlightblue}{$\mathbf{0.057}_{\mathbf{\pm 0.018}}$} & 
            \colorbox{highlightblue}{$\mathbf{0.079}_{\mathbf{\pm 0.034}}$} & 
            \colorbox{highlightblue}{$\mathbf{0.065}_{\mathbf{\pm 0.010}}$} & 
            \colorbox{highlightblue}{$\mathbf{0.042}_{\mathbf{\pm 0.014}}$} \\ 
            
            \bottomrule
        \end{tabular}
    }
    \caption{
        \textbf{Quantitative Results on 3D Action-Conditioned Video Prediction.} We compared our method on 3D action-conditioned video prediction quality with MPM~\cite{hu2018moving}, GBND~\cite{zhang2024dynamic}, and particle-based baselines. The $\mathcal{J}$-Score/IoU and the $\mathcal{F}$-Score measures mask similarities and the LPIPS score measures appearance-wise similarities between predicted frames and ground truth video recordings. We report the mean and standard deviation of the prediction error over a 3-second horizon. The best results are highlighted in bold and blue.
    }
    \label{tab:dynamics_rendering}
    \vspace{-15pt}
\end{table*}

\subsection{Experiment Setup}
We conduct data collection and experiments using a bimanual xArm setup, with each robot arm having seven degrees of freedom. The objects used in the experiments include rope, cloth, plush toys, paper bags, boxes, and bread.

\paragraph{Rope} A single robot arm grasps one end of a rope, while the other end remains free on the table surface. The robot manipulates the rope in 3D space, generating various deformation patterns such as bending and dragging.

\paragraph{Cloth} Two robot arms grasp a rectangular piece of cloth and manipulate it in 3D space. The lower half of the cloth remains in contact with the table, resulting in significant deformations under lifting, moving, and folding actions.

\paragraph{Plush} A single robot arm grasps one limb of a plush toy while the rest of the toy remains in contact with the table. The robot manipulates the plush in 3D space, creating deformation patterns such as limb movements and flipping.

\paragraph{Paper Bag} One robot arm grasps and stabilizes one side of an envelope-shaped mailer bag, while the other manipulates it in 3D space. The robot performs various actions, including opening, closing, and rotating the bag.

\paragraph{Box} Two robot arms are used to open and close shipping boxes. The grippers remain closed, and the manipulation is performed in a nonprehensile manner, utilizing the surfaces of the grippers to push against the movable parts of the box.

\paragraph{Bread} Two robot arms are used to tear pieces of bread. The grippers remain closed, holding the bread in the air. One robot arm stays still while the other pulls, creating stretching effects and eventual breakage.

\begin{figure*}[t]
    \centering
    \includegraphics[width=\linewidth]{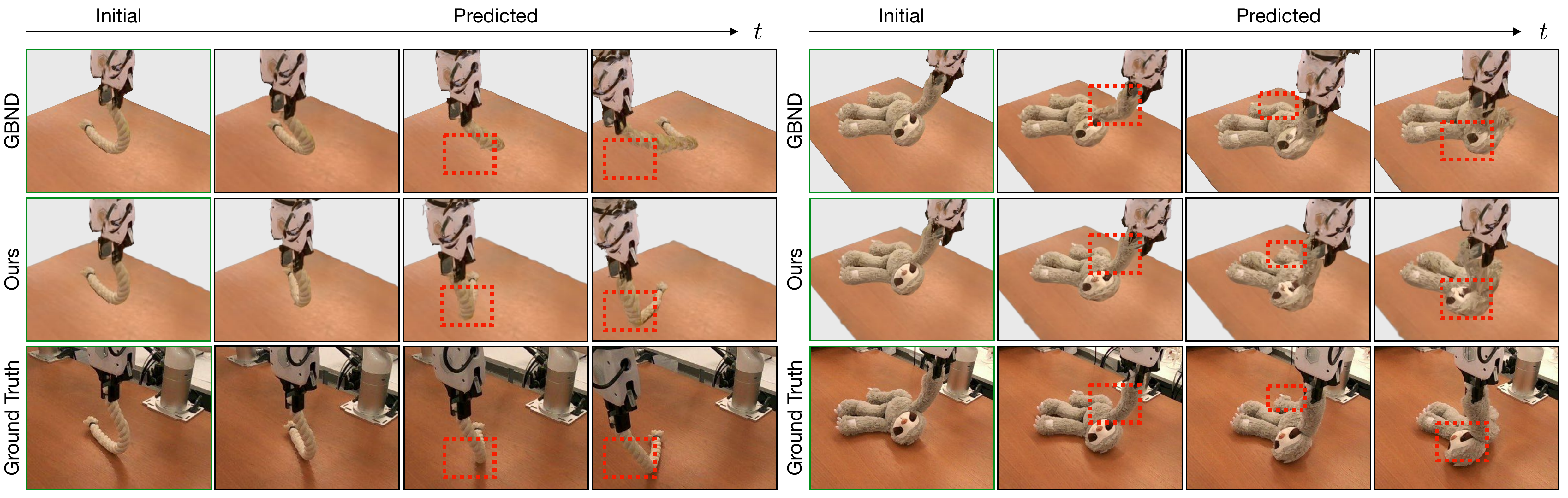}
    \caption{\small
    \textbf{Qualitative Comparisons on 3D Action-Conditioned Video Prediction.} We show our method and the GBND baseline's prediction on two examples of rope and plush toy, compared with the ground truth video. The predictions are based on the 3DGS reconstructions on the first frame (leftmost image) and the robot action sequence. Differences are highlighted with red dashed boxes. Our method aligns better with the ground truth while the baseline method predicts visually nonrealistic deformations.
    }
    \label{fig:qual_video}
    \vspace{-10pt}
\end{figure*}
\begin{figure*}[t]
    \centering
    \includegraphics[width=\linewidth]{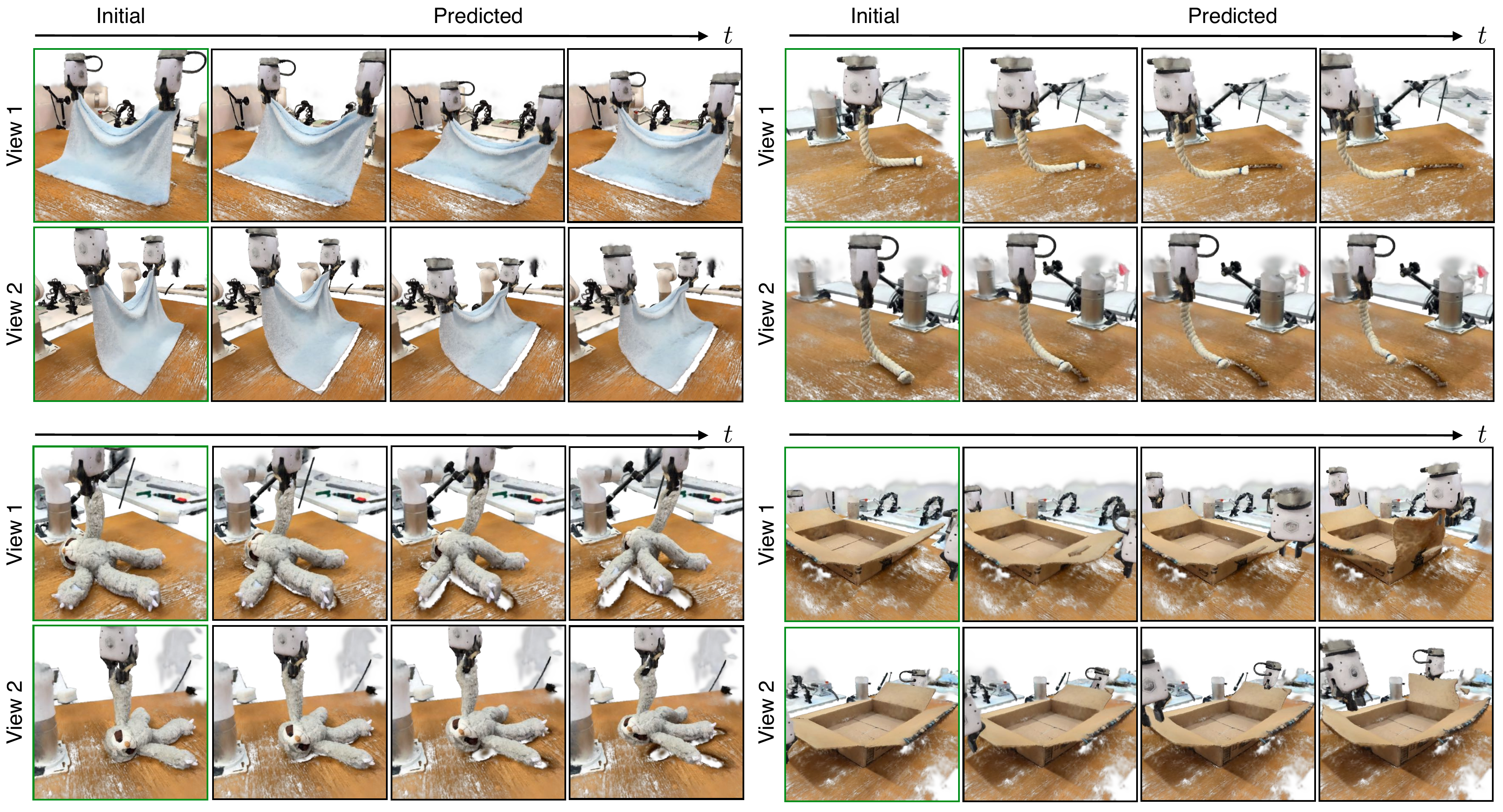}
    \caption{\small
    \textbf{Qualitative Visualizations of Simulation from Scanned Scenes.} Our method gives higher-quality video prediction results with high-resolution Gaussians reconstructed from phone scans. Given the initial reconstruction (green frame), we apply our particle-grid dynamics model to simulate the segmented object, and visualize from different views.
    }
    \label{fig:scanned}
    \vspace{-15pt}
\end{figure*}

The baseline models in our comparisons are as follows:
\begin{itemize}
    \item MPM-based deformable object simulation~\cite{hu2018moving, ma2023learning}: This baseline assumes a hyperelastic material with an unknown uniform Young's modulus and friction coefficient with the tabletop. Parameter identification is performed via gradient descent.
    \item Graph-Based Neural Dynamics (GBND)~\cite{zhang2024dynamic}: This model represents objects using subsampled sparse vertices, along with the robot end-effectors, and employs a Graph Neural Network (GNN) to predict particle motions.
    \item Particle-based dynamics model (ours w/o grid): In this baseline, we ablate the grid representation in our model and directly query the velocity field at particle positions to predict per-particle velocities.
\end{itemize}
For additional information on the experiment setups and baseline implementations, please refer to Appendix~\ref{sec:supp_implement}.

\subsection{Dynamics Learning and Prediction}
We evaluate accuracy using a held-out set of robot-object interactions. The interaction videos are divided into 3-second clips, and the dynamics rollout accuracy is assessed using 3D point cloud metrics that compute the distance between predicted particle positions and ground truth future points. The metrics include Mean Distance Error (MDE, the average distance between the corresponding predicted and ground truth points), Chamfer Distance (CD, the average distance between pairs of nearest neighbors between two point clouds), and Earth Mover’s Distance (EMD, the average distance between two point clouds according to an optimal correspondence). All metrics are calculated with $\SI{}{\meter}$ as the unit of measurement. For the box category, we use the Robot Particles control method representation since it is not directly compatible with MPM. Therefore, we omit MPM from the box comparison.

Quantitative results are shown in Table \ref{tab:dynamics_particle}, where our method outperforms all baselines in terms of dynamics rollout accuracy. We visualize the particle prediction results of our method alongside the baselines in Fig.~\ref{fig:qual_rollout}. Our method's predictions align more closely with the ground truth images and also exhibit higher resolution. In contrast, GBND predicts inadequate particle motions for objects like cloth, plush toy, box, bread, and it generates artifacts for objects like rope. 

\begin{figure*}[t]
    \centering
    \includegraphics[width=\linewidth]{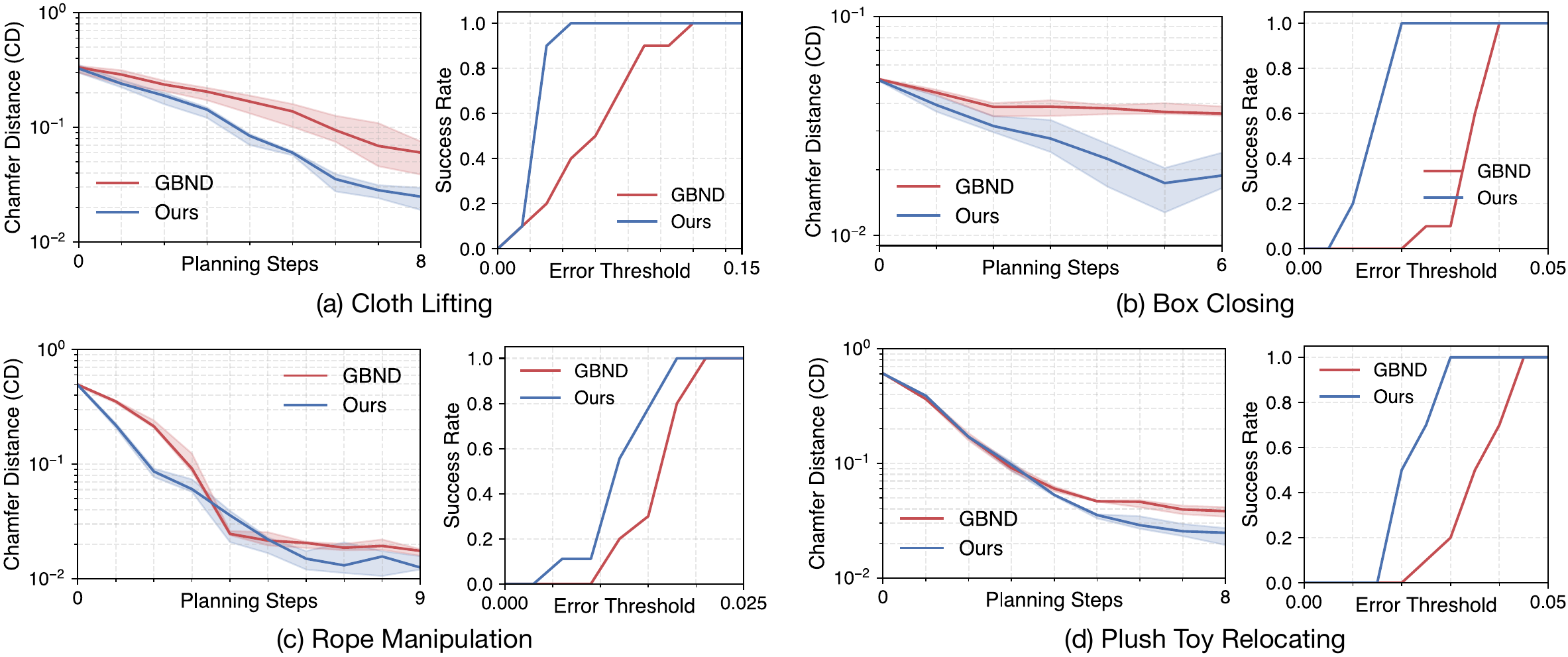}
    \caption{\small
    \textbf{Quantitative Comparisons on Planning.} For four manipulation tasks—cloth lifting, box closing, rope manipulation, and plush toy relocating—we present the error curve and the final success rate curve with respect to the error threshold for task success. The error is always measured using the Chamfer Distance between the current and target point clouds. Our method outperforms the GBND baseline in both error reduction rate and success rate.
    }
    \label{fig:quant_planning}
    \vspace{-10pt}
\end{figure*}
\begin{figure*}[t]
    \centering
    \includegraphics[width=\linewidth]{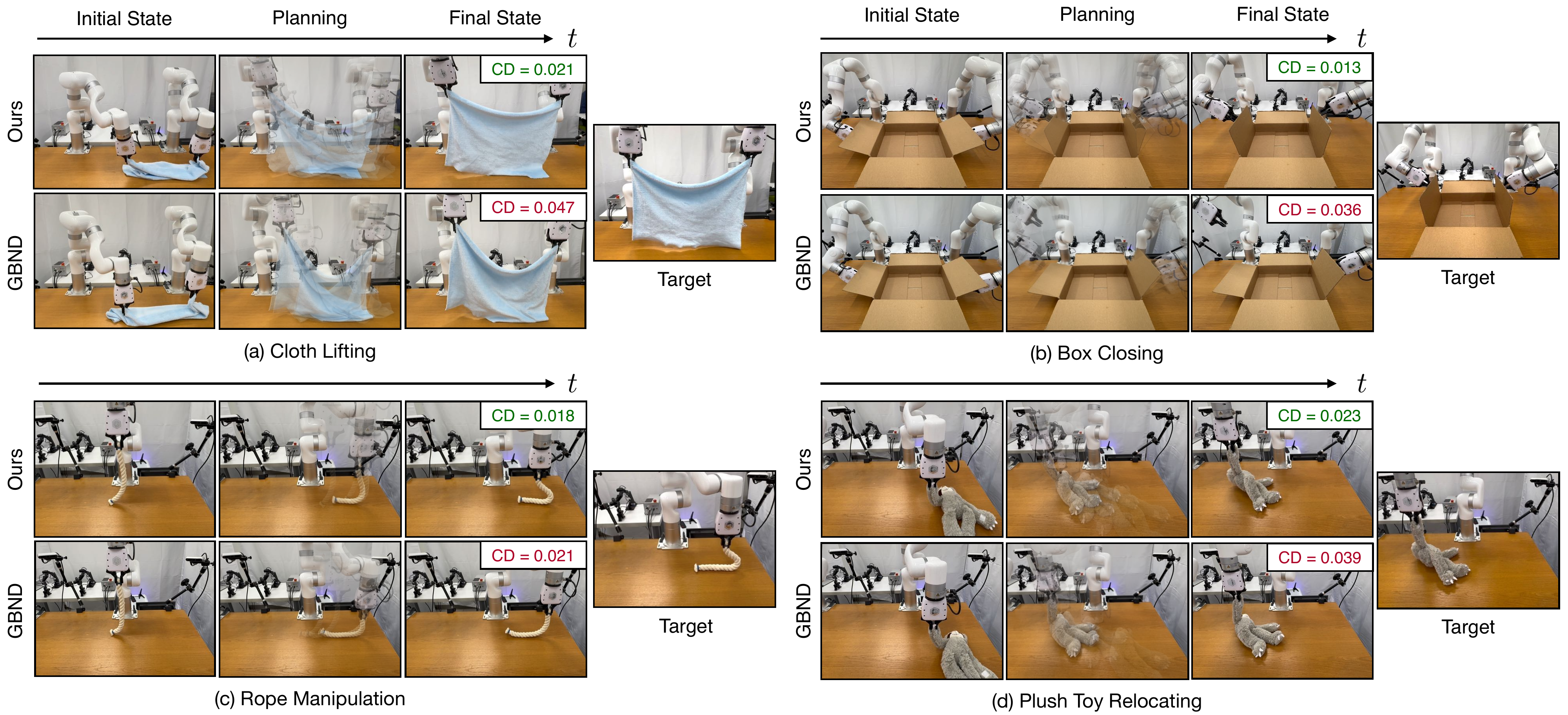}
    \caption{\small
    \textbf{Qualitative Comparisons on Planning.} For each of the four tasks, we visualize a representative planning sequence for both our method and the GBND baseline. Given similar initial states and the same number of planning steps, our method achieves a lower final error, as measured by Chamfer Distance (CD), and produces results that are visually more similar to the target.
    }
    \label{fig:qual_planning}
    \vspace{-15pt}
\end{figure*}

\subsection{Sparse-View Dynamics Prediction}
We evaluate the performance of our method in sparse-view scenarios by training the model on partial observation data. Specifically, during training, we use point cloud from a randomly sampled number of camera views as model input. During evaluation, we test the model's performance with 1 to 4 camera views.

The results shown in Fig.~\ref{fig:partial} demonstrate that our model outperforms the GBND baseline in dynamics prediction accuracy, regardless of the number of input views. Additionally, the performance drop when decreasing the number of views is also less significant than the baseline. For the cloth category, the baseline performance drops significantly when decreasing from 4 camera views to 1 camera view, while our model maintains a low prediction error.

\subsection{Category-Level Model}
Next, we assess the model's ability to generalize across multiple instances within the same category by training on a combined dataset of various object instances. For ropes, the model is trained on 4 distinct ropes and evaluated on 2 unseen ropes. For cloths, it is trained on 6 cloth instances and tested on 2 unseen cloths. The 6 rope instances are cotton rope, jute rope, utility rope, cable, paracord, and yarn, with the cotton rope and utility rope included in the test set and unseen during training. For cloths, the 8 instances include a flannel blanket, cotton towel, microfiber cloth, cotton bed sheet, curtain, wallpaper, mat, and foam sheet, with the flannel blanket and foam sheet included in the test set. These instances are selected to have diverse physical properties and shapes, allowing us to thoroughly evaluate the model's generalization. 

The results in Fig.~\ref{fig:gen} show that our model achieves lower prediction errors than the GBND baseline across both categories and for both seen and unseen instances. Notably, for the cloth category, the baseline method exhibits a significant performance drop on unseen instances, whereas our method keeps a relatively low error, demonstrating better generalization to novel objects at test time.

\subsection{Action-Conditioned Video Prediction}
For action-conditioned video prediction, we use the predicted point cloud trajectories to interpolate Gaussian kernel transformations using LBS~\cite{10.1145/1276377.1276478}. We reconstruct Gaussians from 4 input views using Gaussian Splatting~\cite{kerbl3Dgaussians}. The Gaussians are trained with a segmentation mask loss, following previous works~\cite{zhang2024dynamic, luiten2023dynamic}. Videos are rendered with a fixed input camera pose.
The video prediction quality is assessed using mask-based metrics, including $\mathcal{J}$-Score (IoU), $\mathcal{F}$-Score (contour matching accuracy), and the image-based metric LPIPS.

The resulting metrics are shown in Table \ref{tab:dynamics_rendering}. Our approach achieves the best overall performance. For categories with relatively large objects, for instance, boxes and paper bags, we observe that spiky Gaussian reconstructions often negatively impact mask prediction performance, especially when objects undergo significant deformation. The higher mask alignment scores in GBND and Particle baselines are largely due to inadequate particle motion predictions.

In Fig.~\ref{fig:qual_video}, we show the action-conditioned video prediction results by reconstructing the object using Gaussian Splatting from 4 views and deforming the Gaussians with predictions from our dynamics model. Our method achieves higher-quality rendering and better alignment with the ground truth.

In Fig.~\ref{fig:scanned}, we further demonstrate that our method can be used for simulation based on high-quality phone-scanned GS reconstructions. The scenes are reconstructed using video scans of a static workspace. Coupled with our learned particle-grid neural dynamics, we can generate 3D action-conditioned video predictions with even greater visual fidelity.

\subsection{Planning}
In planning experiments, we evaluate the model's ability to integrate with MPC to generate actions for manipulating objects. We test on 4 tasks with distinct object types: cloth lifting, box closing, rope manipulation, and plush toy relocating. For each task, we conduct 10 repetitive experiments. Performance is assessed using error curves and task success rates.

The quantitative results are shown in Fig.~\ref{fig:quant_planning}. Across all four planning tasks, our method achieves a lower terminal error and a higher error reduction rate compared to the GBND baseline. In Fig.~\ref{fig:qual_planning}, we visualize the initial states, intermediate steps, and final states, comparing them to the target. In all four tasks, our method produces results that are visually closer to the target.
For example, in the box closing task, our method successfully lifts both sides of the box, whereas the baseline struggles to predict the correct actions and often loses contact with the box. In rope manipulation, our method accurately bends the rope by pressing downward, while the baseline fails to achieve this due to lower prediction resolution.

\section{Limitations}
While we have demonstrated that Particle-Grid Neural Dynamics can model diverse types of deformable objects, the current framework has several limitations:
(i) The current formulation assumes a fixed number of particles during iterative rollout, making it inapplicable to scenarios involving the appearance or disappearance of particles. This limitation could be addressed by correcting the particle sets with new observations or modeling the per-frame visibility of particles.
(ii) The model implicitly infers an object's physical properties from its point cloud and short-term motion history. While this is sufficient for modeling a single object instance or multiple instances with distinct shapes and physical properties, a more systematic approach to modeling physical properties is needed for interpretable identification and adaptation at test time. This could involve learning a parameter-conditioned neural dynamics model~\cite{zhang2024adaptigraph}.
(iii) Training the model and applying it to video prediction depend on accurate predictions from computer vision models such as Segment-Anything~\cite{kirillov2023segment}, CoTracker~\cite{karaev24cotracker3}, and Gaussian Splatting~\cite{kerbl3Dgaussians}. Failures in these perception and reconstruction models could negatively impact our method’s performance.

\section{Conclusion}
\label{sec:conclusion}
In this paper, we introduce Particle-Grid Neural Dynamics, a novel framework for learning neural dynamics models of deformable objects directly from sparse-view RGB-D recordings of robot-object interactions. By leveraging a hybrid particle-grid representation to capture object states and robot actions in 3D space, our method outperforms previous graph-based neural dynamics models in terms of prediction accuracy and modeling density. This advancement enables the modeling of a wide range of challenging deformable objects.
Additionally, integration with 3D Gaussian Splatting facilitates 3D action-conditioned video prediction, simultaneously capturing both object geometry and appearance changes, thereby creating a learning-based digital twin of real-world objects. We further demonstrate that our model can be applied to various deformable object manipulation tasks, achieving improvements in both task execution efficiency and success rate.

\section*{Acknowledgment}
We thank the members of the RoboPIL lab at Columbia University for their helpful discussions. This work is partially supported by the Toyota Research Institute (TRI), the Sony Group Corporation, Google, and Dalus AI.
This article solely reflects the opinions and conclusions of its authors and should not be interpreted as necessarily representing the official policies, either expressed or implied, of the sponsors.

\bibliographystyle{plainnat}
\bibliography{refs}

\clearpage
\begin{center}
\textbf{\textsc{\Large Appendix}}
\end{center}

\newcommand\DoToC{%
    \hypersetup{linkcolor=black}
  \startcontents
  \printcontents{}{1}{\textbf{Contents}\vskip3pt\hrule\vskip3pt}
  \vskip7pt\hrule\vskip3pt
}

\begin{appendices}

\renewcommand{\thesubsection}{\Alph{section}.\arabic{subsection}}

\begin{figure*}[t]
    \centering
    \includegraphics[width=\linewidth]{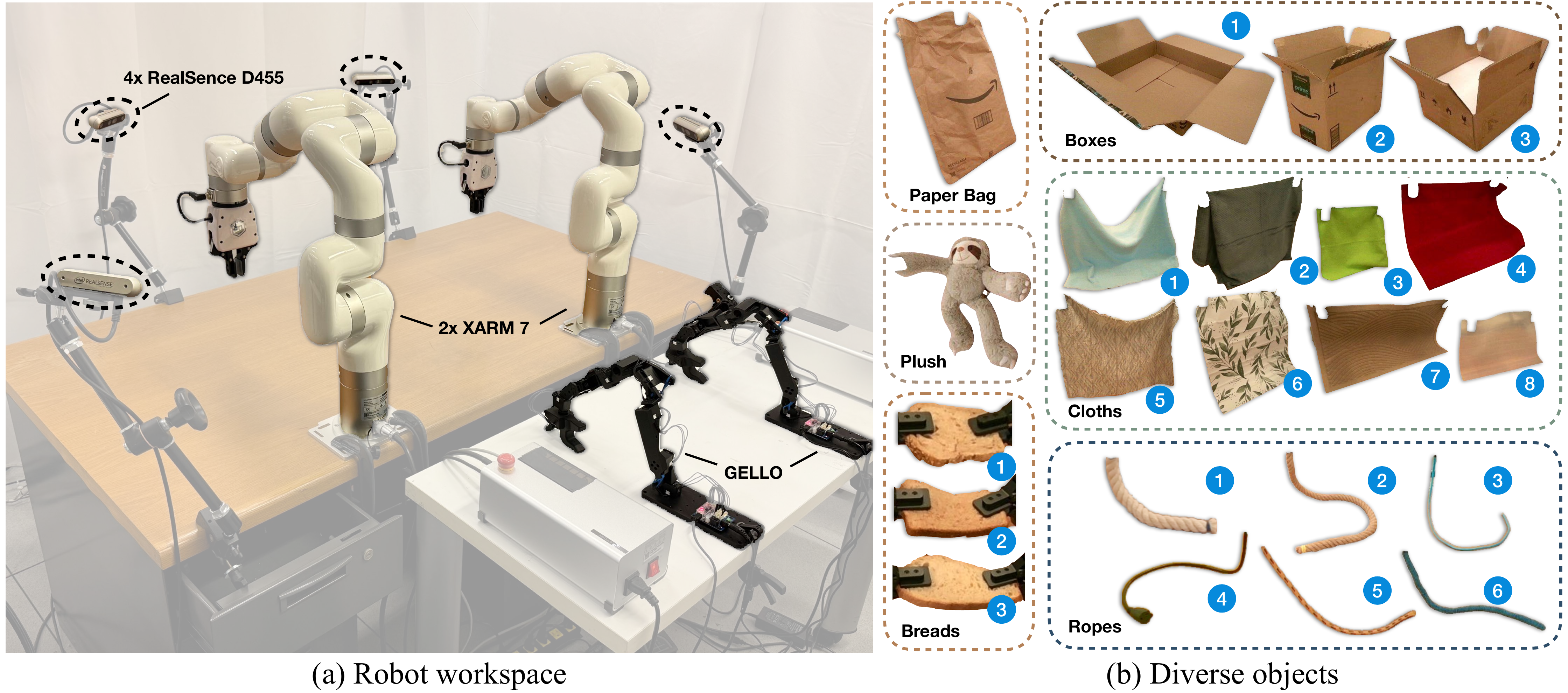}
    \caption{\small
    \textbf{Experiment Setup.} 
    \textbf{(a)} Our robot workspace includes four calibrated RGB-D cameras positioned at each corner of the table, along with a GELLO~\cite{wu2023gello} system for teleoperating the dual xArm 7 robotic arms equipped with parallel grippers.
    \textbf{(b)} Our experiments involved six types of materials: (i) a paper bag, (ii) a stuffed animal, (iii) three varieties of bread, (iv) three boxes of different shapes, (v) eight cloth pieces varying in fabric type and size, and (vi) six ropes differing in length, thickness, and stiffness.
    }
    \label{fig:exp_setup}
    \vspace{-15pt}
\end{figure*}

\vspace{5pt}
\DoToC
\vspace{10pt}

\hypersetup{linkcolor=red}
\section{Method Details}

\subsection{Data Collection}
\label{sec:supp:data_collect}
In this work, all training data are collected through an object segmentation and tracking pipeline that utilizes foundational vision models. The pipeline consists of the following steps: 
\begin{itemize}
    \item Object detection and segmentation;
    \item 2D tracking and 3D velocity computation;
    \item 3D tracking extraction.
\end{itemize}

\paragraph{Object detection and segmentation}
or each camera view, we apply Grounded-SAM-2~\cite{ravi2024sam2segmentimages, kirillov2023segment, ren2024grounded} to extract object masks. Using text prompts containing object descriptions, Grounded-SAM-2 detects and segments object in the first frame of a sequence and then propagates the mask across subsequent frames. To apply the model, we partition the recording into non-overlapping clips, each containing 720 frames (i.e., 24 seconds).

\paragraph{2D tracking and 3D velocity computing}
We use CoTracker~\cite{karaev24cotracker3} as the tracking model due to its stable and robust performance in 2D tracking. Each camera view's recording is parsed into clips of 30 frames (i.e., 1 second) with 15 overlapping frames. Within each clip, we crop objects from the video based on the segmentation mask and fill the background with black pixels. Using the cropped the clip as input, CoTracker predicts a set of 2D trajectories, initialized from uniformly sampled grid locations in the first frame of the clip. We then interpolate each pixel's 3D velocity using the predicted trajectories and the corresponding depth image, as described by the following equation:
\begin{equation}
    \mathbf{v}_{i, t} = \frac{1}{\Delta t} \left( \text{proj}^{-1}_{t+\Delta t} (\hat{x}_{i,t+\Delta t}) - \text{proj}^{-1}_{t} x_{i,t}\right),
\end{equation}
where $\text{proj}^{-1}_{t}(\cdot)$ is the inverse projection function that maps a pixel location to its corresponding 3D point using depth and camera transformation. $x_{i, t}$ denotes the 2D location of pixel $i$ at time $t$, and $\hat{x}_{i,t+\Delta t}$ is the interpolated location of pixel $i$ at the next time step, $t+\Delta t$. The resulting $\mathbf{v}_{i, t}$ represents the 3D velocity of pixel $i$ in the world frame at time $t$. 

To ensure that each pixel has sufficient nearby trajectories for interpolation, we restrict velocity prediction to the first 15 frames of the clip. With the 15 overlapping frames between adjacent clips, we ensure that each frame contains velocity predictions. Finally, we combine the per-pixel velocities from all camera views to compute the point cloud velocity for each frame.

\paragraph{3D tracking extraction}
After obtaining the 3D point cloud with per-point velocities, we extract persistent point tracks uusing an iterative rollout approach. Starting from frame $t$, we iteratively evolve the particles using the current velocity $\mathbf{v}$. At each subsequent time step, we perform $k$-NN to identify the closest points and update the particle velocities to the mean velocity of the selected neighbors.

In summary, our segmentation and tracking pipeline does not rely on differentiable rendering or optimization, making it computationally more efficient. Moreover, since foundation models typically perform well on various kinds of texture-less deformable objects, such as cloth and bread, our tracking model is also capable of handling these challenging objects.

\subsection{Video Prediction}
\label{sec:supp_video_pred}
In this section, we detail how we interpolate the 6DoF transformation of the Gaussian kernels based on particle motion predictions. Given the reconstructed Gaussian $\mathcal{G} = \{\mathbf{X}_{\text{GS}}, \mathbf{C}, \mathbf{R}_{\text{GS}}, \mathbf{S}, \mathbf{O}\}$, the predictions $\hat{\mathbf{X}}_{t+\Delta t}$, and the previous-frame particle positions $\mathbf{X}_t$, we first calculate the rotations of each particles, assuming local rigidity over a small time gap. For each particle $p$, we identify its $k_{\text{rot}}$ nearest particles, denoted as $\mathcal{N}(p)$. We then calculate the rotation $\mathbf{r}_p$ of particle $p$ by
\begin{equation}
    \mathbf{r}_p = \arg \min_{\mathbf{r} \in \text{SO}(3)}\sum_{s\in \mathcal{N}(p)} \|\mathbf{r} (\hat{\mathbf{x}}_{s, t+\Delta t} - \hat{\mathbf{x}}_{p, t+\Delta t}) - (\mathbf{x}_{s, t} - \mathbf{x}_{p, t})\|^2.
\end{equation}
Equivalently, $\mathbf{r}_p$ represents the optimal rotation that transforms the nearest particles of $p$ to their new positions. Using the estimated rotations, we perform Linear Blend Skinning (LBS)~\cite{10.1145/1276377.1276478} to interpolate the transformations. For a Gaussian kernel $i$ with $\mu_{i, t} \in \mathbf{X}_{\text{GS}, t}$ and $q_{i, t} \in \mathbf{R}_{\text{GS}, t}$, we denote its $k_{\text{LBS}}$ nearest particles to be $\mathcal{N}(i)$. We calculate the interpolated positions and quaternions, $\hat{\mu}_{i, t+\Delta t}$ and $\hat{q}_{i, t+\Delta t}$, using the following equations:
\begin{gather}
    \hat{\mu}_{i, t+\Delta t} = \sum_{p\in \mathcal{N}(i)} w_{ip} \mathbf{r}_p (\mu_{i, t} - \mathbf{x}_{p, t}) + \mathbf{x}_{p, t} + (\hat{\mathbf{x}}_{p, t+\Delta t} - \mathbf{x}_{p, t}), \label{eq:lbs_1}\\
    \hat{q}_{i, t+\Delta t} = \left(\sum_{p\in \mathcal{N}(i)} w_{ip, t} \mathbf{r}_{p, t}\right) \odot q_{i, t}, \label{eq:lbs_2}
\end{gather}
where $\odot$ denotes quaternion multiplication, and $w_{ip, t}$ is the interpolation weight between Gaussian kernel $i$ and particle $p$ at time $t$, calculated as:
\begin{equation}
    w_{ib, t} = \frac{\|\mu_{i, t} - \mathbf{x}_{p, t}\|^{-1}}{\sum_{s\in \mathcal{N}(i)} \|\mu_{i, t} - \mathbf{x}_{s, t}\|^{-1}}.
\end{equation}
Intuitively, this weight function assigns larger weights to the particles that are spatially closer to the Gaussian kernel. The weights are used to compute the weighted average of particle rotations and translations, which are then applied to the kernel (as shown in Eq. \ref{eq:lbs_1} and \ref{eq:lbs_2}). Empirically, we set the $k$-NN parameters for rotation and LBS to $k_{\text{rot}} = k_{\text{LBS}} = 8$.

\subsection{Model-Based Planning}
\label{sec:supp_planning}
For the four manipulation experiments, the planning settings are detailed as follows. The cloth lifting and box closing tasks are bimanual, while the rope manipulation and plush toy relocation tasks involve only a single arm. For the latter two tasks, we do not consider gripper rotations, so the action space consists of the 3-DoF $(x, y, z)$ position of the end effector in the task space. For the two bimanual tasks, we account for 3D rotations, resulting in an action space that includes the translation and rotation of both grippers, totaling 12 degrees of freedom. In each iteration of the MPPI~\cite{williams2017model} algorithm, we sample $N$ delta actions from a normal distribution: 
\begin{equation}
    \Delta \mathbf{x}_{\text{eef}}^{(i)}, \Delta \mathbf{r}_{\text{eef}}^{(i)} \sim N(\mathbf{0}, \Sigma), \ \  i=1,\cdots, N,
\end{equation}
where $\Delta \mathbf{x}_{\text{eef}}$ and $\Delta \mathbf{r}_{\text{eef}}$ represent the residual translations and rotations of the end-effector. We then construct action trajectories by adding the delta actions to the current optimal actions:
\begin{align}
    \{(\mathbf{x}_{\text{eef}, 1:T}, \mathbf{r}_{\text{eef}, 1:T}) | &\mathbf{x}_{\text{eef}, 1:t} = \mathbf{x}_{\text{eef}}^* + t\cdot \Delta \mathbf{x}_{\text{eef}}^{(i)}, \\
    &\mathbf{r}_{\text{eef}, 1:t} = \mathbf{r}_{\text{eef}}^* + t\cdot \Delta \mathbf{r}_{\text{eef}}^{(i)}, \\
    & t=1, \cdots, N\}.
\end{align}
We then evaluate the sampled action trajectories using the learned dynamics model and calculate the cost based on Chamfer Distance (Eq.~\ref{eq:planning}). For the cloth lifting task, to prevent damage to the cloth, we include an additional penalty term with an indicator function that penalizes trajectories where the distance between the grippers exceeds $L=\SI{0.6}{\meter}$.

\section{Implementation Details}
\label{sec:supp_implement}

\subsection{Experiment Setup}
\label{sec:supp_exp_setup}

Our robot setup and experimental objects are shown in Fig.~\ref{fig:exp_setup}. 
Our experiments, including real-world data collection and manipulation tasks, are conducted in a workspace (Fig.~\ref{fig:exp_setup}a) equipped with four calibrated RealSense D455 cameras. These cameras are positioned at the corners of the table to capture RGB-D images at 30Hz with a resolution of 848x480. Additionally, we have integrated the GELLO~\cite{wu2023gello} teleoperation system with dual UFACTORY xArm 7 robotic arms, each with 7 degrees of freedom and xArm’s parallel grippers, enabling a human operator to collect bi-manual random interaction data.

We consider 6 types of deformable objects in our work (Fig.~\ref{fig:exp_setup}b): (i) a paper bag, (ii) a stuffed animal plush toy, (iii) three types of bread, (iv) 3 boxes, (v) 8 cloth pieces differing in fabric, texture, and size, and (vi) 6 ropes varying in length, thickness, and stiffness. For the dynamics prediction experiments (Table \ref{tab:dynamics_particle} and \ref{tab:dynamics_rendering}), we use only one type of cloth and rope. The remaining objects are used only in the category-level model experiments (Fig.~\ref{fig:gen}).

\subsection{Baselines Details}
\label{sec:supp_baselines}

In this section, we provide additional details on the implementation of the baseline methods, which include both physics-based and learning-based simulators for deformable objects, as well as an ablated version of our method that excludes the grid representation.

\subsubsection{MPM}
\label{sec:supp_baselines_mpm}
We follow previous works~\cite{xie2023physgaussian, ma2023learning} for the implementation and parameter optimization of MPM. In the MPM algorithm, an object is discretized into Lagrangian particles that carry dynamic information such as position, velocity, strain, and stress. The physical properties of the object are described by the material's elasticity and plasticity models, known as constitutive models. Different object types require different constitutive models, and objects within the same type may have varying parameters. We assume that the materials considered in our work are hyperelastic, and thus we use the fixed corotated elasticity model with an identity plasticity model. The optimizable parameters in the models include Young's modulus $E$, which characterizes the stiffness or stress in response to deformations, and the Poisson's ratio $\nu$, which is kept fixed. We also optimize the friction coefficient $\mu$ between the object and surface. Since we assume uniform material properties across the objects, the optimizable parameters $E$ and $\mu$ are shared across all particles.

We optimize the MPM model using the same training setting as our particle-grid neural dynamics model. Since it is infeasible to identify per-frame internal strain in the objects, we assume that the object is in its rest state with no strain at the first frame of each data sequence. The MPM model is trained using the same loss function (Eq.~\ref{eq:loss}). We employ Adam as the optimizer with gradient clipping and perform gradient descent optimization on both parameters.

\subsubsection{GBND}
\label{sec:supp_baselines_gbnd}
The GBND method defines the environment state as a graph: $z_t \overset{\underset{\mathrm{\Delta }}{}}{=} \GG_t =  ( \VV_t, \EE_t )$, where $\VV_t$ is the vertex set representing object particles, and $\EE_t$ denotes the edge set representing interacting particle pairs at time step $t$. 
Given dense particles of object and robot end-effector pose, we apply farthest point sampling~\cite{FarthestPointSampling} to downsample object into sparse vertices, while representing the robot gripper with a single particle. 
For a vertex $v_{i, t} \in \VV_t$, we incorporate the history positions $\mathbf{x}_{i, t-h\Delta t:t}$ to implicitly capture velocity information, along with vertex attributes $\mathbf{c}_{i, t}$ to indicate whether the particle corresponds to the object or the robot end-effector. The final input is the concatenated feature vector $(\mathbf{x}_{i, t-h\Delta t:t}, \mathbf{c}_{i, t})$. In practice, we set $h=2$ for all materials. For end-effector particles, the action $\mathbf{u}_t$, represented as delta gripper transformation, is also included in the input feature. The edges between particles are dynamically constructed over time by identifying the nearest neighbors of each particle within a predefined radius. Additionally, we impose a limit on the maximum number of edges that any single node can have, which we empirically set to 5.

After constructing the graph, we instantiate the dynamics model as graph neural networks (GNNs) that predict the evolution of the graph representation $z_t$. The graph design follows previous works~\cite{li2018learning, zhang2024dynamic}. To control the accumulation of dynamics prediction errors, we supervise the model's predictions over $K=4$ steps. Formally, we train the model using following the loss function:

\begin{gather}
    \mathcal{L}_{\text{train}} = \frac{1}{|\mathcal{V}|}\sum_{i=1}^{|\mathcal{V}_0|}\sum_{t=1}^K \|\hat{\mathbf{x}}_{i,t} - \mathbf{x}_{i,t}\|^2 + \\
    \lambda \cdot \frac{1}{|\mathcal{E}|}\sum_{(i, j)\in \mathcal{E}}\sum_{t=1}^{K-1}
    \left(
    \|\hat{\mathbf{x}}_{i, t+1} - \hat{\mathbf{x}}_{j,t+1} \|
    -
    \|\hat{\mathbf{x}}_{i, t} - \hat{\mathbf{x}}_{j,t} \|
    \right)^2
\end{gather}
where $\hat{\mathbf{x}}_{i,t}$ and $\mathbf{x}_{i,t}$ represent the predicted and ground truth positions of the $i^{\text{th}}$ vertex, respectively. The first term in the loss is the mean squared error, while the second term is an edge length regularization term, which has been shown to be effective for objects with complex shapes~\cite{zhang2024dynamic}. Empirically, we set $\lambda = 0.1$.

\subsubsection{Particle}
\label{sec:supp_baselines_particle}
In this baseline, we ablate the grid representation in our model and directly query the velocity field at particle positions to predict per-particle velocities. Specifically, we predict the particle velocities with a modified version of Eq.~\ref{eq:query}:
\begin{equation}
    \hat{\mathbf{v}}_{p, t+\Delta t} = \mathbf{f}^{\text{field}}_\psi (\gamma(\mathbf{x}_{p}), \bar{\mathbf{z}}_{p, t}),
\end{equation}
where $\bar{\mathbf{z}}_{p, t}$ is the locally-averaged particle features calculated similarly to $\mathbf{z}_{g, t}$ in Eq.~\ref{eq:local_feature}. After calculating $\hat{\mathbf{v}}_{p, t+\Delta t}$, we apply a similar process as in Grid Velocity Editing to edit particle velocities according to grasping actions and collisions with the table. The remaining parts of the model, as well as the training of the Particle baseline, are identical to those in our model. 

By simplifying the grid representation, the velocity field must give predictions for the entire set of particles. Compared to querying the field at fixed grid locations, this increases the difficulty of learning the velocity field MLP. Since the input coordinates are processed with high-frequency sinusoidal positional encoding $\gamma$, the spatial smoothness of the field is not regularized. Empirically, we show that this negatively impacts the dynamics prediction accuracy in the experiments results in the main paper (Table \ref{tab:dynamics_particle} and \ref{tab:dynamics_rendering}).
\begin{figure*}[t]
    \centering
    \includegraphics[width=\linewidth]{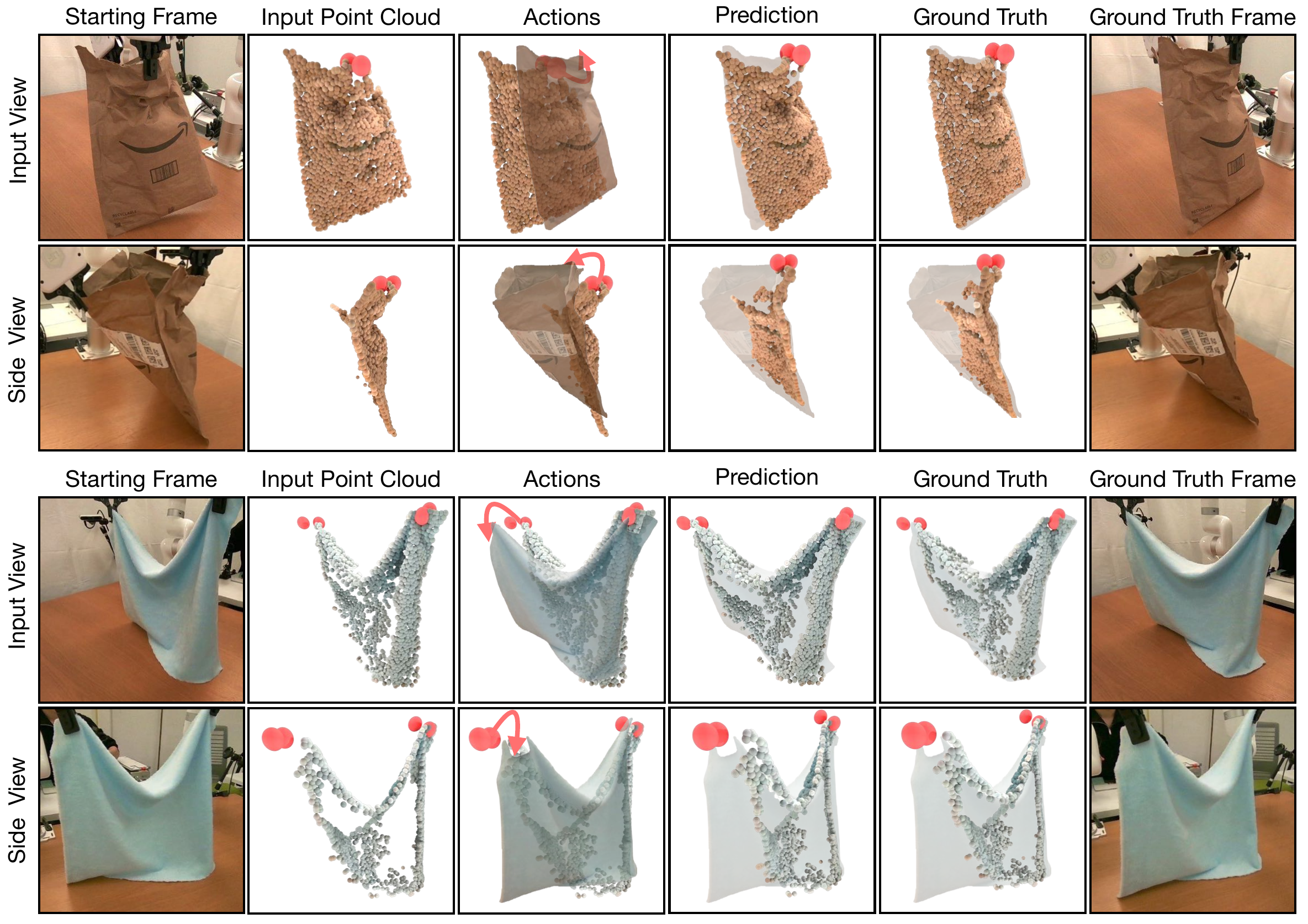}
    \caption{\small
    \textbf{Additional Qualitative Comparisons on Partial Observation.} In this experiment, we use only point clouds from a single view as input to the model. We visualize the starting frame and the input point clouds, followed by our prediction results compared to the ground truth future particle positions and the future frame. Additionally, we provide a side view where the incompleteness of the input point cloud is more apparent. Our model's predictions closely align with the ground truth, demonstrating its robustness under partial observations.
    }
    \label{fig:supp_partial}
    \vspace{-10pt}
\end{figure*}
\begin{figure*}[t]
    \centering
    \includegraphics[width=\linewidth]{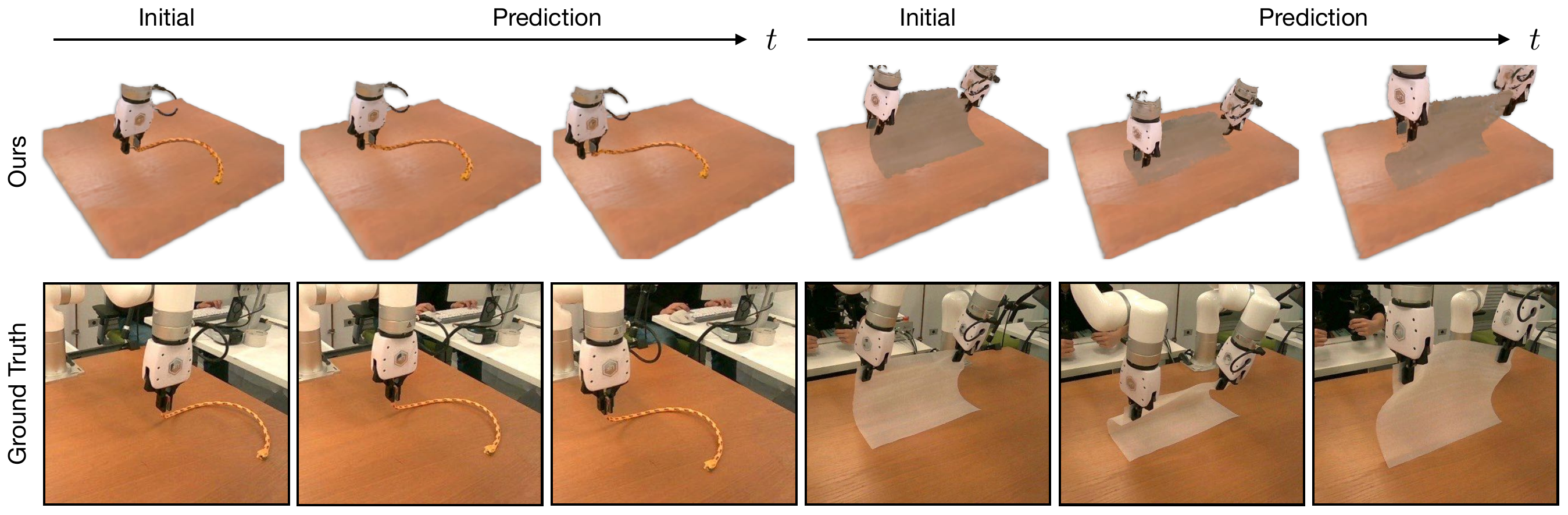}
    \caption{\small
    \textbf{Additional Qualitative Comparisons on Generalization.} In this experiment, we deploy the trained model on objects not seen during training. We visualize the video prediction results using Gaussian Splatting, showing both the starting and predicted frames. The predicted outcomes closely match the ground truth, demonstrating the model's ability to generalize to unseen instances.
    }
    \label{fig:supp_gen}
    \vspace{-5pt}
\end{figure*}

\subsection{Method Implementation}
\label{sec:supp_method_implement}
Following previous works~\cite{xie2023physgaussian, ma2023learning}, we implement the grid velocity editing and G2P processes with NVIDIA Warp~\cite{warp2022}, which accelerates the differentiable process on GPU. Training our model typically takes around 8 hours on a single NVIDIA 4090 GPU, using a batch size of 32. 

In all experiments, we use a time step of $\Delta t = \SI{0.1}{\second}$. To effectively model smaller objects, such as bread, we introduce a scaling factor $s$ that scales the object before being input into the model. Specifically, we use $s=3.0$ and grid size $\delta=\SI{1}{\centi\meter}$ for bread, $s=0.8, \delta=\SI{2}{\centi\meter}$ for cloth, and $s=1.0, \delta=\SI{2}{\centi\meter}$ for all other object categories. In all experiments, the particle velocities are set to zero at the start of an episode, assuming that the object starts from a static state particle. Otherwise, the velocities are computed from the difference between current and previous observations.

The training dataset size, measured by the total duration of recorded videos, is as follows: (i) cloth: 20.3 minutes, (ii) rope: 21.7 minutes, (iii) plush toy: 3.8 minutes, (iv) box: 10.1 minutes, (v) paper bag: 6.7 minutes, and (vi) bread: 4.8 minutes. The evaluation dataset for each category contains test episodes of each 3 seconds in length, as this duration captures challenging deformations while remaining computationally efficient. We use 40 test episodes for cloth and rope, and 20 test episodes for all other categories.

For the friction coefficient parameter in ground contact, we fix it empirically to reduce complexity, as our experiments do not involve extreme friction variations and the fixed value yields strong performance across tasks.

\section{Additional Experiments}

\subsection{Qualitative Comparisons on Partial Views}
In Fig.~\ref{fig:supp_partial}, we provide qualitative visualizations of our model's prediction when using partial view inputs. Specifically, we visualize the input partial view from a single depth image, alongside our model's predictions and the ground truth. The results demonstrate that our model is capable of making accurate predictions using despite being given only a highly incomplete set of particle data.

\subsection{Qualitative Comparisons on Generalization}
In Fig.~\ref{fig:supp_gen}, we present qualitative visualizations of our model's predictions on unseen object instances. Specifically, we show video predictions on unseen cloth and rope instances and compare them against the ground truth. The results demonstrate that our model generalizes well to novel objects not encountered during training and predicts realistic rendering results.

\subsection{Qualitative Comparisons on Dynamics Prediction}
\label{sec:supp_qual_dyn}
\begin{figure*}[t]
    \centering
    \includegraphics[width=\linewidth]{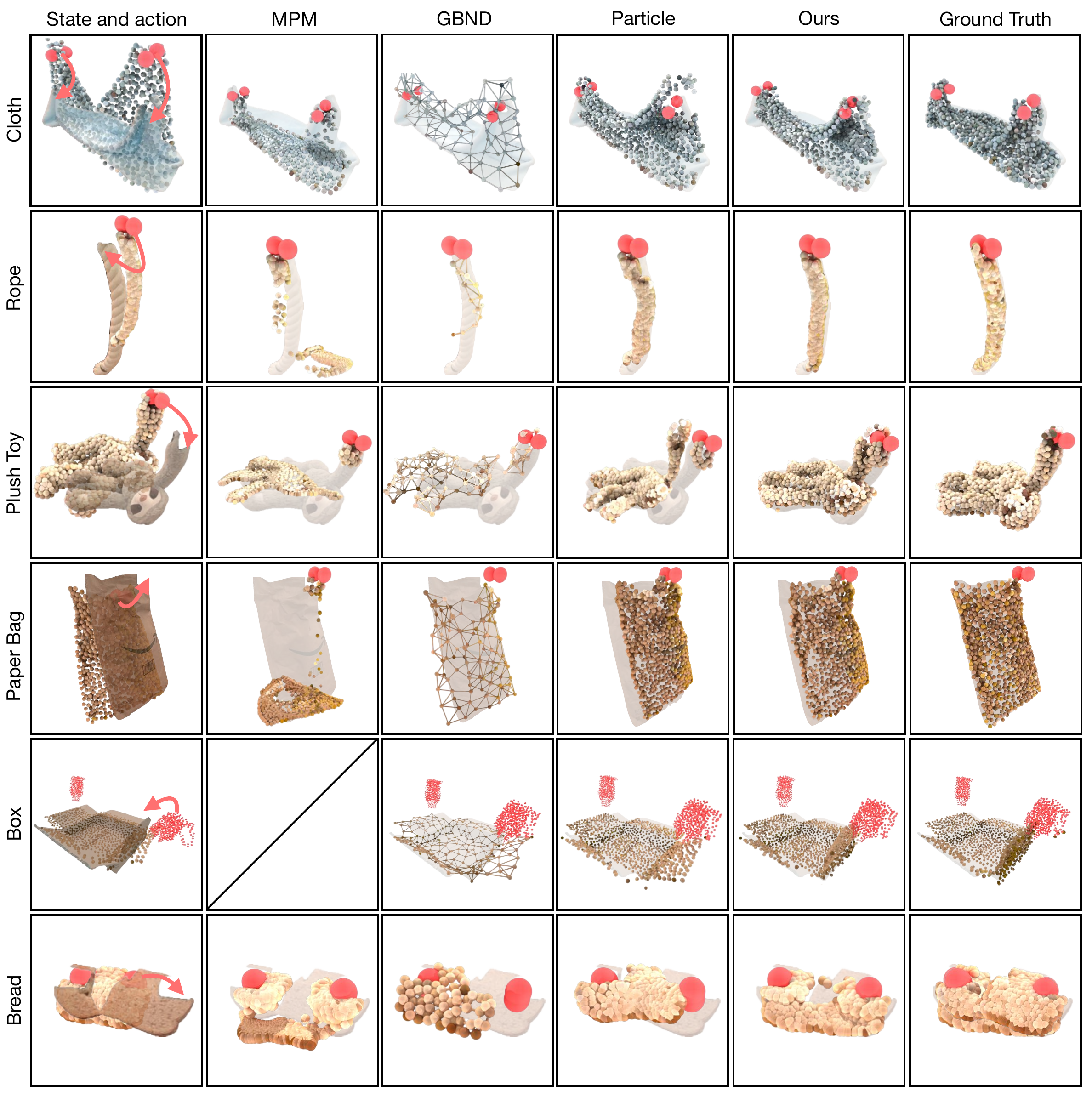}
    \caption{\small
    \textbf{Additional Qualitative Comparisons on Dynamics Prediction.} Given the initial states and actions (leftmost column), we present the prediction results of the MPM with parameter identification baseline, the GBND baseline, and the Particle baseline, compared to our particle-grid neural dynamics model (middle columns), compared to the ground truth particles at the final frame (rightmost column). We use a pair of red spheres to indicate the position and orientation of robot grippers. The half-transparent background in the middle columns are the ground truth final state images, which help highlight the prediction errors. From the results, we observe that our method's predictions align the best with the ground truth, with fewer artifacts. In contrast, the baseline methods tend to predict motions that do not match well with the ground truth (e.g., GBND-paper bag, MPM-cloth), cause unwanted breakage (e.g., MPM-rope, GBND-plush toy), or display insufficient particle motions (e.g. GBND-box, Particle-box, Particle-bread).
    }
    \label{fig:supp_rollout_1}
    \vspace{-15pt}
\end{figure*}

\begin{figure*}[t]
    \centering
    \includegraphics[width=\linewidth]{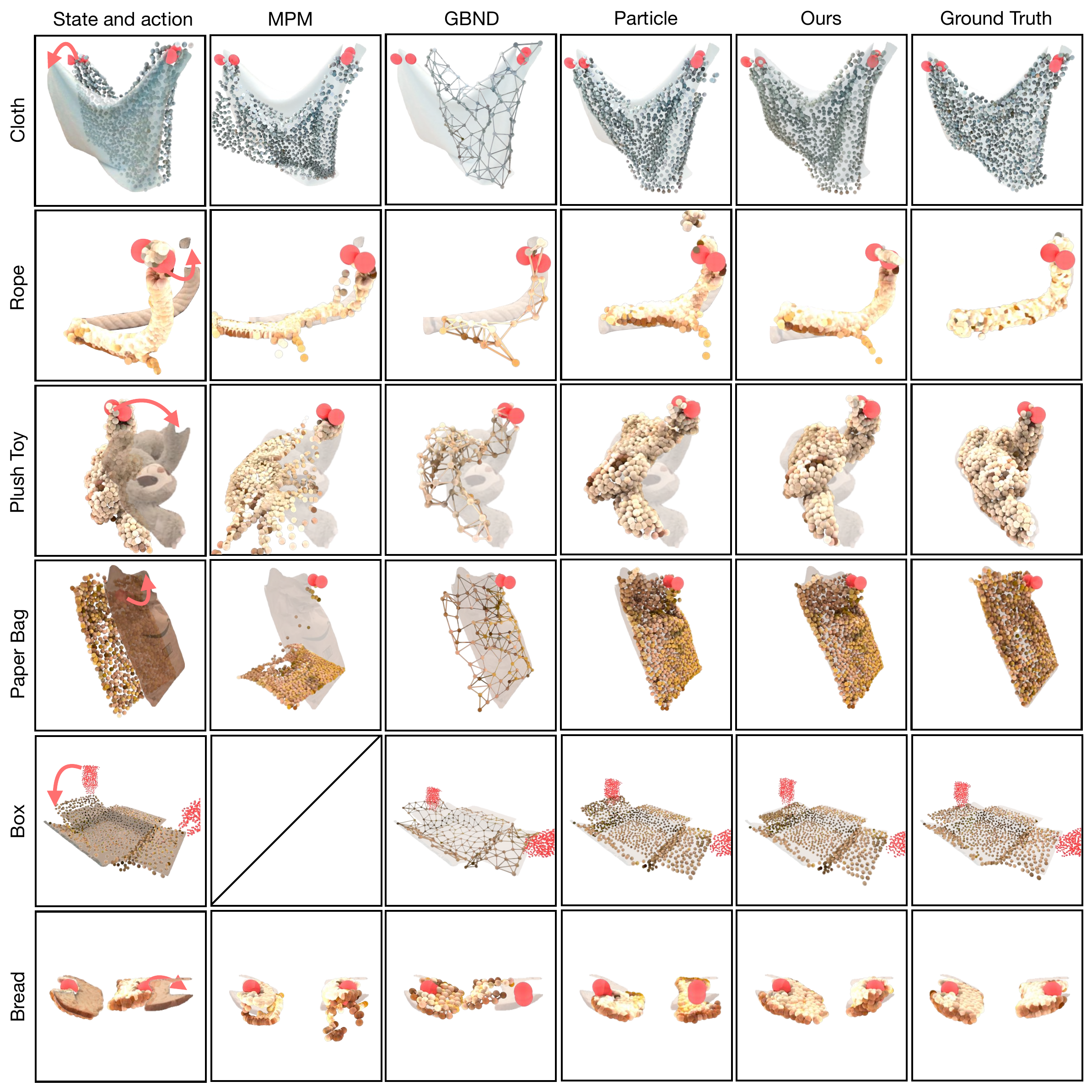}
    \caption{\small
    \textbf{Additional Qualitative Comparisons on Dynamics Prediction.} Given the initial states and actions (leftmost column), we show the prediction results of the MPM with parameter identification baseline, the GBND baseline, and the Particle baseline, compared to our particle-grid neural dynamics model (middle columns), compared to the ground truth particles at the final frame (rightmost column). We use a pair of red spheres to indicate the position and orientation of robot grippers. The half-transparent background in the middle columns are the ground truth final state images, which help highlight the prediction errors. From the results, we observe that our method's predictions align the best with the ground truth, with fewer artifacts. In contrast, the baseline methods tend to predict motions that do not match well with the ground truth (e.g., GBND-plush toy, MPM-cloth), cause unwanted breakage (e.g., MPM-paper bag, MPM-bread), or display insufficient particle motions (e.g. Particle-plush toy, Particle-box).
    }
    \label{fig:supp_rollout_2}
    \vspace{-15pt}
\end{figure*}

In Fig.~\ref{fig:supp_rollout_1} and \ref{fig:supp_rollout_2}, we provide additional qualitative comparisons between our method and the three baselines on dynamics prediction. These results extend those presented in Fig.~\ref{fig:qual_rollout} to include the MPM and Particle baselines. From the qualitative examples, we can observe that the MPM baseline tends to predict soft, drifting motions that do not align well with the ground truth. Furthermore, the particles in the MPM model tend to scatter or fall to the ground, likely due to the insufficient point density in the interior volume and the effects of gravity. The Particle-only baseline predicts dense particle motions similar to our method. However, it predicts inadequate particle motions and creates artifacts near the gripper, especially for objects like plush toys, boxes, and breads.


\subsection{Additional Quantitative Experiments}
\begin{table}[t]
\centering
\begin{tabular}{@{}lcccc@{}}
\toprule
Batch size & 1 & 10 & 50 & 100 \\ \midrule
Time ($\SI{}{\ms}$) & 4.8 & 5.2 & 11.9 & 22.3 \\ \bottomrule
\end{tabular}
\caption{\textbf{Runtime analysis.}}
\label{tab:runtime}
\end{table}

\begin{table}[t]
\centering
\begin{tabular}{@{}lccc@{}}
\toprule
Method & MDE & CD & EMD \\ \midrule
Cloth GVE & \textbf{0.045} & \textbf{0.043} & \textbf{0.022} \\
Cloth RP & 0.058 & 0.051 & 0.026 \\ 
Box GVE & 0.027 & 0.035 & 0.019 \\
Box RP & \textbf{0.022} & \textbf{0.015} & \textbf{0.016} \\ \bottomrule
\end{tabular}
\caption{\textbf{Ablation study on deformation controlling methods.}}
\label{tab:deform}
\vspace{-10pt}
\end{table}

\begin{table}[t]
\centering
\begin{tabular}{@{}lccc@{}}
\toprule
Method & MDE & CD & EMD \\ \midrule
Ours-1.25$\SI{}{\cm}$ & \textbf{0.039} & \textbf{0.037} & \textbf{0.021} \\
Ours-4$\SI{}{\cm}$ & 0.051 & 0.049 & 0.028 \\ 
MLP Encoder & 0.067 & 0.068 & 0.044 \\
PTv3 Encoder~\cite{wu2024point} & \underline{0.040} & 0.039 & \underline{0.022} \\ 
Ours & \textbf{0.039} & \underline{0.038} & \textbf{0.021} \\ \bottomrule
\end{tabular}
\caption{\textbf{Ablation study on grid size and point encoder design.}}
\label{tab:ablation}
\end{table}

\begin{table}[t]
\centering
\begin{tabular}{@{}lccc@{}}
\toprule
Method & Seen/Unseen MDE & 1-4 Views MDE  & Planning CD \\ \midrule
MPM & 0.188/0.195 & 0.186/0.186/0.179/0.177 & 0.030$_{\pm \text{0.010}}$ \\
Ours & \textbf{0.047}/\textbf{0.056} & \textbf{0.059}/\textbf{0.058}/\textbf{0.056}/\textbf{0.053} & \textbf{0.011}$_{\pm \text{\textbf{0.003}}}$ \\ \bottomrule
\end{tabular}
\caption{\textbf{Further comparison with MPM.}}
\label{tab:mpm}
\vspace{-10pt}
\end{table}

\subsubsection{Runtime Analysis}
We evaluate our model’s runtime on the rope test set, with results shown in Table~\ref{tab:runtime}. During inference, our method runs at $\SI{4.8}{\milli\second}$ per forward pass with a batch size of 1, and at $\SI{11.9}{\milli\second}$ with a batch size of 50, enabling real-time, batched inference crucial for planning.

\subsubsection{Ablation Studies}
To validate our design choice of using different deformation control methods for different action types, we compare the two approaches in Table~\ref{tab:deform}. Results show that Grid Velocity Editing (GVE) performs better for prehensile tasks (e.g., cloth), while Robot Particles (RP) are more effective for nonprehensile cases (e.g., boxes), supporting the claims made in Sec.~\ref{sec:GridVelocityEditing}.
Furthermore, to assess the choice of PointNet~\cite{qi2017pointnet} as our point encoder and the grid size configuration, we perform ablation studies on the grid size (1.25$\SI{}{\cm}$ and 4$\SI{}{\cm}$ vs. our default 2$\SI{}{\cm}$) and the encoder design (shared MLP and PTv3~\cite{wu2024point} vs. PointNet). Results on the rope dataset, shown in Table~\ref{tab:ablation}, indicate that while a grid size of 1.25$\SI{}{\cm}$ and PTv3 achieve performance comparable to ours, they incur higher computational costs.

\subsubsection{Further Comparison with MPM}
Additional comparisons with the model-based simulator MPM on rope data are provided in Table~\ref{tab:mpm}. Empirically, we find that despite having explicit material parameters, MPM underperforms our method across all tasks due to its inability to handle partial observations from depth inputs.

\end{appendices}

\end{document}